\documentclass[lettersize,journal]{IEEEtran}

\usepackage{amssymb,amsmath}
\usepackage{cite}
\usepackage{graphicx}
\usepackage{subfig}

\usepackage{psfrag}
\usepackage{url}
\usepackage[latin1]{inputenc}
\usepackage[absolute,overlay]{textpos}
\usepackage{gensymb}
\usepackage{cases}
\usepackage[font=footnotesize]{caption}
\usepackage{float}
\usepackage[linesnumbered,ruled,lined]{algorithm2e}
\usepackage{pgf}

\usepackage[nocomma]{optidef}

\usepackage{amsthm}

\usepackage{booktabs}
\usepackage{color,soul}

\newcommand{\algo}[1]{\textbf{\algorithmcfname~\ref{#1}}} 

\usepackage{textcomp}
\usepackage{bbm}
\usepackage{xcolor}
\def\BibTeX{{\rm B\kern-.05em{\sc i\kern-.025em b}\kern-.08em
    T\kern-.1667em\lower.7ex\hbox{E}\kern-.125emX}}

\usepackage{tabularx}
\usepackage{makecell}
\usepackage{multirow}

\begin{document}

\title{Diffusion Offline Reinforcement Learning for Fair and Energy-Efficient UAV-Assisted Wireless Networks}
\author{
	\IEEEauthorblockN{Eslam Eldeeb and Hirley Alves
}
	
    \thanks{Eslam Eldeeb and Hirley Alves are with the Centre for Wireless Communications (CWC), University of Oulu, Finland. (e-mail: eslam.eldeeb@oulu.fi; hirley.alves@oulu.fi).
    }
    
    \thanks{This work was supported by 6G Flagship (Grant Number 369116) funded by the Research Council of Finland, and supported by the Business Finland project, 6G-FISRE.}
}
\maketitle

\begin{abstract}
The integration of generative artificial intelligence with wireless communication and signal processing systems has opened new avenues for intelligent, data-driven decision-making in future 6G networks. This work proposes a diffusion soft actor-critic (Diffusion-SAC) approach that leverages offline reinforcement learning (RL) enhanced by denoising diffusion probabilistic models (DDPMs) to optimize trajectory and scheduling control in unmanned aerial vehicle (UAV) networks. While offline RL methods, such as conservative Q-learning (CQL), can learn from static datasets, they often struggle to generalize in low-data or dynamic conditions. To address this, we combine the robustness of CQL with the generative power of diffusion models, enabling expressive and signal-aware policy learning that generalizes beyond behavior policies. Applied to a UAV-assisted wireless network, the proposed framework minimizes transmission energy and improves fairness among devices. Simulations show that Diffusion-SAC outperforms standard offline RL baselines, achieving more stable convergence and higher rewards even with limited datasets. The method enhances data efficiency, reduces energy consumption, and increases throughput by more than $35 \%$ compared to existing algorithms, demonstrating its potential for robust policy learning in next-generation wireless control systems.
\end{abstract}
\begin{IEEEkeywords}
	Diffusion models, offline reinforcement learning, soft actor-critic, energy minimization, unmanned aerial vehicles
\end{IEEEkeywords}

\vspace{-1mm}

\section{Introduction}\label{sec:introduction}

Recent advances in generative modeling have enabled expressive data-driven representations for complex signal processing and control tasks in future wireless networks~\cite{10515203,11098641}. Among these models, diffusion models (DMs) have demonstrated strong capabilities in modeling high-dimensional structured distributions through iterative denoising processes. In wireless communications, generative models have shown promise in applications such as channel estimation and prediction~\cite{9252921}, beamforming and resource allocation in massive MIMO systems~\cite{10422716}, digital twin modeling~\cite{10628026}, and UAV-assisted networking~\cite{wang2024energy}.

Despite their expressive power, purely generative models remain inherently data-driven and typically rely on imitation of observed behaviors without explicitly optimizing long-term objectives~\cite{cao2025reinforcementlearninggenerativeai}. In dynamic, decision-centric wireless environments, where channel conditions, traffic demands, and energy constraints continuously evolve, effective control policies must adapt beyond the behavior observed in static datasets \cite{10529221}. For example, in unmanned aerial vehicle (UAV)-assisted networks, mobility decisions must account for time-varying channel conditions, device requirements, and energy constraints. UAVs provide flexible relay capabilities by repositioning to serve distant or low-power devices, thereby improving link quality and enhancing communication reliability through adaptive spatial signal processing~\cite{xia2024standoff}.

In such UAV-assisted wireless networks, trajectory planning and signal-aware mobility control are critical components of efficient system design. UAV mobility introduces an additional degree of freedom that can be exploited to optimize key performance metrics, including coverage, latency, energy efficiency, and information freshness~\cite{8870206}. However, determining optimal UAV paths is challenging due to dynamic wireless environments, heterogeneous quality-of-service (QoS) requirements, and strict energy constraints. Moreover, UAVs often operate under limited environmental knowledge, necessitating algorithms that are both data-efficient and generalize to unseen scenarios~\cite{9933838}. These challenges motivate the development of intelligent decision-making frameworks that learn effective movement strategies from historical data while maintaining adaptability during deployment~\cite{eldeeb2022multi,9507262}.

To address the need for goal-directed improvement, reinforcement learning (RL) provides a mathematically grounded framework for sequential decision-making~\cite{sutton1999reinforcement}. An RL agent interacts with an environment by observing states, selecting actions, and receiving rewards, with the objective of maximizing cumulative long-term returns. In its traditional online form, RL enables exploration-driven policy refinement and can discover strategies that surpass those observed in training data, making it particularly suitable for adaptive signal processing and wireless control tasks.

However, online RL can be costly, time-consuming, and potentially unsafe in resource-constrained domains such as UAV-assisted wireless systems. Furthermore, purely online training disregards valuable offline datasets that may be available from prior deployments or simulation logs. Offline RL addresses this limitation by enabling policy learning entirely from static datasets, without requiring live interaction with the environment~\cite{levine2020offline}. By decoupling learning from active exploration, offline RL facilitates safer and more practical deployment in wireless signal processing and control systems~\cite{10753476}.

Nevertheless, naively applying standard RL algorithms to offline data often leads to distributional shift and extrapolation errors~\cite{levine2020offline}. Conservative offline RL methods, such as conservative Q-learning (CQL) and implicit Q-learning (IQL), mitigate these issues by constraining value estimation and reducing over-optimism~\cite{eldeeb2025offlinedistributionalreinforcementlearning}. While these approaches enhance stability and safety, their performance may still be limited by restrictive policy parameterizations when modeling complex or structured action distributions. In addition, their effectiveness depends on the richness and diversity of the offline dataset. Diffusion models offer an expressive alternative for policy representation. Their iterative denoising mechanism enables flexible modeling of complex action distributions over continuous or hybrid spaces, making them promising candidates for structured wireless control problems.

In this work, we propose a novel offline RL framework, named Diffusion-SAC, that combines the generative capabilities of diffusion models with the robustness of conservative Q-learning for adaptive signal processing and control in wireless UAV networks. Our contributions are summarized as follows:
\begin{itemize}
    \item We introduce a diffusion-based policy parameterization for hybrid discrete-continuous offline RL, enabling joint UAV trajectory and scheduling optimization under realistic wireless constraints.

    \item We provide a systematic analysis of the interaction between behavior cloning, Q-guidance, and conservative regularization, revealing that diffusion models can act as implicit distributional regularizers, reducing the reliance on explicit CQL penalties in offline signal processing control.
    
    \item We demonstrate that diffusion-based policies exhibit robust generalization under limited and imperfect datasets, and we quantify the effects of dataset size, dataset quality, and denoising depth on convergence stability and energy-AoI trade-offs.

    \item Simulation results demonstrate the superiority of our method over conventional SAC and behavior cloning baselines. An ablation study further illustrates the benefits of diffusion policy modeling in sparse-data scenarios.
\end{itemize}
The remainder of the paper is organized as follows: Section~\ref{LitRev} reviews related literature. Section~\ref{sec:system_model} introduces the system model and problem formulation. Section~\ref{sec:backg} provides the necessary background on SAC, CQL, and diffusion models. Section~\ref{sec:ODRL} details our proposed Diffusion-SAC algorithm. Numerical evaluations are provided in Section~\ref{sec:results}, and Section~\ref{sec:conclusions} concludes the paper.

\vspace{-2mm}

\section{Related Work}\label{LitRev}
In this section, we provide an overview of the recent progress in offline RL, diffusion models, and diffusion RL in solving signal processing and wireless communications-related problems.

\vspace{-2mm}

\subsection{Offline Reinforcement Learning}

Offline RL, also known as batch RL, is a subdomain of RL in which agents learn optimal policies offline, without interacting with the environment. Offline RL addresses concerns in online RL, such as unsafe and costly exploration. Training RL agents using offline, pre-collected datasets often introduces distributional shift and estimation bias due to the discrepancy between the learned policies and the behavior policies used to collect the datasets. Famous methods, including CQL~\cite{kumar2020conservative} and IQL~\cite{kostrikov2021offlinereinforcementlearningimplicit}, solve this problem by explicitly penalizing the Q-values of the out-of-distribution (OOD) actions and implicitly constraining the Q-function to the observed dataset, respectively. On the other hand, classifier-free diffusion guidance has been used to narrow the offline-to-online distribution gap, yielding consistent gains during fine-tuning~\cite{Huang2025CFDG}.

Offline RL has been introduced recently to the wireless domain as an efficient and safe decision-maker. The authors of~\cite{10636835} introduce offline RL with CQL to enable efficient computation offloading for AR tasks over Terahertz wireless networks. The work in~\cite{10529190} evaluates offline RL algorithms by mixing offline datasets for radio resource management optimization, while the authors in~\cite{eldeeb2025resilientuavtrajectoryplanning} present a resilient meta-offline RL approach for UAV trajectory planning that enhances spatial signal processing and energy efficiency.

While conservative offline RL algorithms, such as CQL, have shown promise in mitigating distributional shift and overestimation issues, they often rely on simplistic policy parameterizations that may be insufficient in complex or high-dimensional spaces. In particular, modeling complex or highly structured action distributions with conventional architectures can lead to suboptimal policies that fail to fully exploit the richness of the dataset. Additionally, offline datasets often contain suboptimal or biased trajectories, making it difficult for the agent to distinguish between high-quality and low-quality actions. These challenges lead to instability and poor generalization, particularly in high-dimensional or sparse-reward settings.

\vspace{-2mm}
\subsection{Diffusion Models}

Diffusion models have emerged as powerful generative signal-processing tools for image, video, and text generation~\cite{orig_diff}. Inspired by non-equilibrium thermodynamics, they outperform other generative models, such as generative adversarial networks (GANs), by learning structured denoising mappings from random noise to high-quality data representations. The denoising diffusion probabilistic model (DDPM)~\cite{NEURIPS2020_4c5bcfec} is a popular diffusion model originally designed for image generation. It includes forward and reverse processes. In the former, stochastic perturbations (noise) are gradually added to the data until it becomes Gaussian, whereas the latter uses a learning model to recover the original data from the noise. Relying on this capability, the model can generate data from pure noise. 

Diffusion models have emerged as key enablers in next-generation wireless signal processing and communications. To this end, the authors of~\cite {10480348} propose diffusion models to mitigate wireless channel noise in image semantic communications. The work in~\cite{10896580} designs a latent diffusion model that performs few-step low-latency denoising over noisy channels. In~\cite{10437154}, diffusion models are used for wireless channel modeling and sampling, while the authors of~\cite {10599525} propose an adaptive modulation and coding scheme using latency-aware diffusion models for extremely low-data-rate semantic communication. Authors in~\cite{Wang2025RISDiff} adopt diffusion to optimize a large-scale 3D-reflective intelligent surface (RIS) deployment. A recent survey consolidates diffusion models for wireless networks, including channel modeling, integrated sensing and communications (ISAC), resource management, and RL-based control, underlining their growing role in statistical signal processing and optimization~\cite{Luong2025Survey}.

\vspace{-2mm}
\subsection{Diffusion Reinforcement Learning}

Recently, diffusion models have gained further attention in the RL domain. Diffusion Q-learning (DQL) is a breakthrough in diffusion RL, where the authors propose a policy regularization using behavior cloning for offline RL via a conditional diffusion model. Later, diffusion models have been used to completely express actors in actor-critic algorithms and generate policies~\cite{wang2023diffusionpoliciesexpressivepolicy}.

Adopting diffusion RL to the wireless domain has been limited. The work in~\cite{10409284} proposed a diffusion  RL framework for AI-generated content services. Similarly, the authors of~\cite {10900473} propose a diffusion-based online RL algorithm to enhance cooperative offloading and resource allocation, while the authors of~\cite {10978657} exploit diffusion models with online RL for radio resource management in radio access network (RAN) slicing. Additionally, \cite{Nouri2025DiffusionRL} explores a diffusion-RL scheme for resource blocks and power allocation, reporting improvements over deep Q-network (DQN) and semi-supervised variational autoencoder baselines. In contrast to existing diffusion-RL approaches that primarily target online RL settings or purely continuous control domains, our work focuses on offline hybrid control for wireless signal optimization, where safety, dataset bias, and distributional shift are critical concerns.

\begin{figure}[t!]
    \centering    \includegraphics[width=0.9\columnwidth,trim={0 0 0 0},clip]{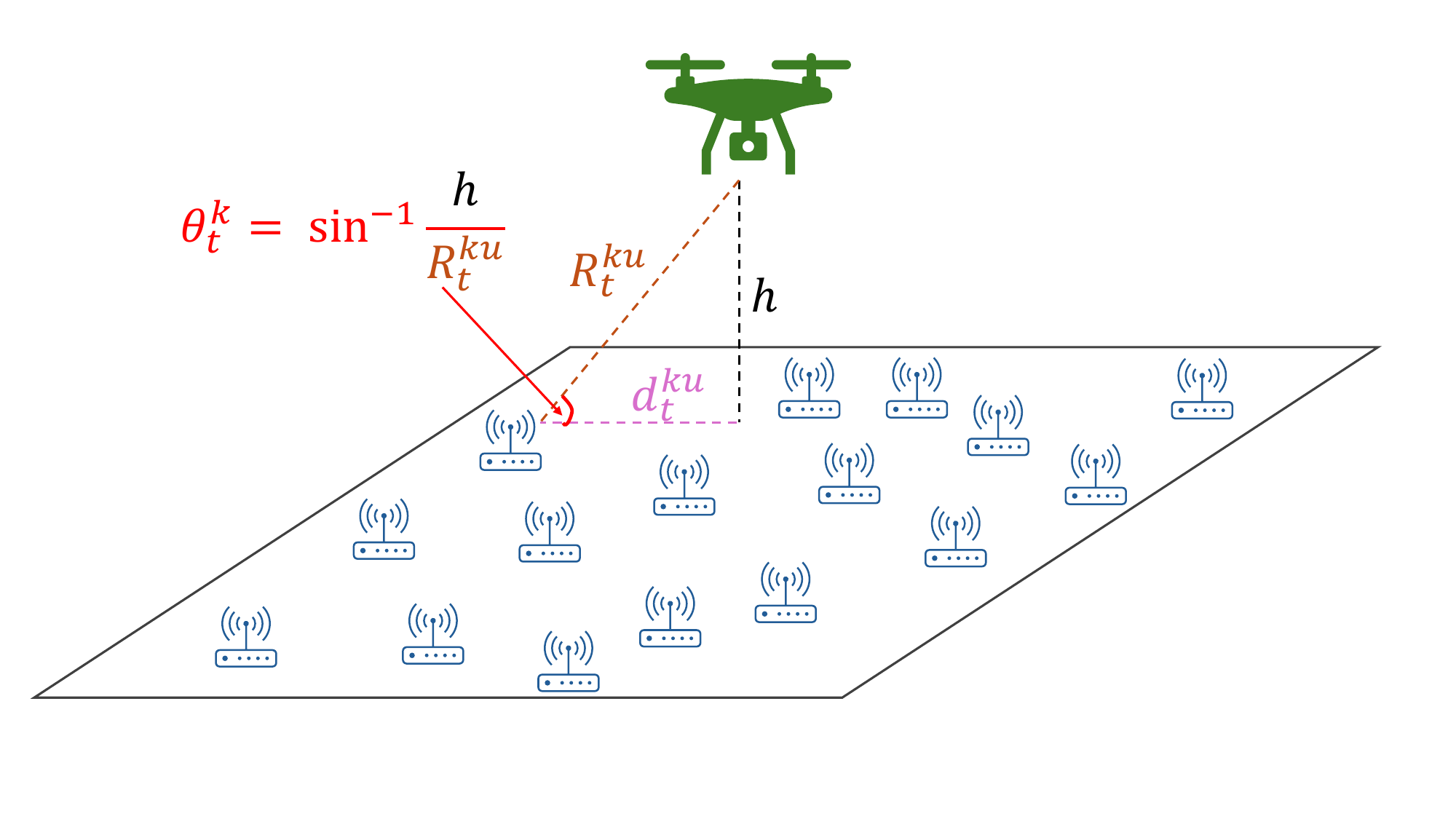} 
    \vspace{-2mm}
    \caption{UAV-IoT system model. A single UAV moves at altitude $h$ while serving $K$ ground devices located at $\{(x_k,y_k)\}_{k=1}^K$. At time $t$, the UAV position is $(x_t^u,y_t^u)$, the link distance to device $k$ is $R_{ku}^t$, and the elevation angle is $\theta_k^t$. Links consist of small-scale fading and a path loss that may be LoS/NLoS with probability dependent on $\theta_k^t$. The corresponding path-loss and rate models are used for scheduling and energy accounting. Each device maintains an AoI state $A_k^t$ with optional fairness weights $\delta_k$. The control action selects a continuous UAV motion and a discrete transmission decision (one of the $K$ devices or idle).}
    \vspace{-2mm}
    \label{UAV_Model}
\end{figure}

\section{System Model}\label{sec:system_model}

We consider a 3D control environment, as illustrated in Fig.~\ref{UAV_Model}, where $K$ fixed, limited-power devices are served by a rotary-wing UAV flying at a fixed altitude $h$ and with a horizontal speed of $v^u$ meters per second. The coordinates of each device $k$ are denoted by $(x^k, y^k)$, while the projected 2D coordinates of the UAV at time $t$ are given by $(x^u_t, y^u_t)$. At each decision step, the UAV selects one device to receive data from. The duration of each time step varies with the quality of the wireless channel, propagation characteristics, and the UAV's current movement decision, as explained below.

The objective is to jointly optimize the UAV's trajectory and scheduling decisions. Specifically, the UAV must sequentially choose its movement direction (continuous action) and the device to schedule for communication $s_t \in {1, \dots, K}$ (discrete action). We model the problem as an episodic environment: each episode begins with the UAV in a randomly initialized position and terminates after a fixed duration, thereby capturing the temporal evolution of spatial signal conditions and control actions within a constrained flight window.

\subsection{Channel Model}
We consider both line-of-sight (LoS) and non-line-of-sight (NLoS) ground-to-air wireless propagation links between the devices and the UAV. The probability of having LoS or NLoS communication between device $k$ and the UAV at time $t$ is, respectively, \cite{8870206}:
\begin{align}
    \label{LoS_Prob}
    p^{ku}_{t} &= \left(1+C \exp\left(- D (\theta^k_t - C) \right)\right)^{-1} \quad \mathrm{if~LoS}, \\
    \bar{p}^{ku}_t &= 1- p^{ku}_t \quad \mathrm{if~NLoS}. 
\end{align}
where $C$ and $D$ are environment-specific constants, and $\theta^k_t$ is the elevation angle between device $k$ and the UAV at time $t$ computed as $\theta^k_t = \arcsin\left(\frac{h}{R^{ku}_t}\right)$, where $R$ is the Euclidean distance between device $k$ and the UAV at time $t$ given by $R^{ku}_t = \sqrt{h^2 + d^2}$, where $d=d_t^{ku} = \sqrt{(x^k-x^u_t)^2 + (y^k-y^u_t)^2}$ denotes the horizontal distance between device $k$ and the UAV at time $t$. Following~\cite{8708295,9892691}, the path loss for LoS and NLoS scenarios is formulated as:
\begin{align}
    \label{Path_loss_models}
    &\Xi_t^{ku}(\xi) = 20 \log_{10} \left( \frac{4 \pi f_c R_t^{ku}}{v^{l}}\right) + \xi \quad \mathrm{[dB]},
\end{align}
where $f_c$ is the carrier frequency, $v^l$ is the speed of light, and $\xi$ is the path loss that can be either $\xi^{\text{LoS}}$ or $\xi^{\text{NLoS}}$ for LoS and NLoS scenarios, respectively. Hence, the average path loss between device $k$ and the UAV at time $t$ is computed as:
\begin{equation}
    \label{avg_path_loss}
    \bar{L}^{ku}_t = p^{ku}_t \: \Xi_t^{ku}(\xi^{\text{LoS}}) + \bar{p}^{ku}_t \: \Xi_t^{ku}(\xi^{\text{NLoS}}) \quad \mathrm{[dB]}.
\end{equation}

Using the average path loss, the average transmission rate in bits per second (bps) for device $k$ at time $t$ is calculated as:
\begin{equation}
    \label{rate}
    r_t^k = B \: \log_2(1+\gamma_t^k) \quad \mathrm{[bps]},
\end{equation}
where $B$ is the available bandwidth in Hertz (Hz) and $\gamma_t^k$ is the signal-to-noise ratio (SNR) given by $\gamma_t^k = 
|h_t^{ku}|^2 \bar{l}^{ku}_t 
\, \frac{P}{\sigma^2}$, where $\bar{l}^{ku}_t = 10^{{-\bar{L}^{ku}_t}/{10}}$, 
$h_t^{ku}$ is the small-scale fading, $P$ is the device transmission power and $\sigma^2 = N_0B$ is the noise power with power spectral density (PSD) $N_0$.

\subsection{Energy Calculation}
The total energy consumption of the UAV is decomposed into communication-related (transmission and signal processing) and movement-related components. The communication-related energy consumption at time step $t$ is given by:
\begin{equation}
    \label{Energy_comm}
    E_t^{\text{com}} = T_{\text{com}}^k (P_{\text{hover}} + P_{\text{com}}) \quad \mathrm{[J]},
\end{equation}
where $T_{\text{com}}^k = {D_k}/{r^k_t}$ is the hovering time of the UAV needed for a device $k$ to complete its data transmission, $D_k$ is the packet size in bits, $P_{\text{com}}$ is the communication power of the UAV, and $P_{\text{hover}}$ is the hovering power calculated as~\cite{8613833}:
\begin{equation}
    \label{hover_power}
    P_{\text{hover}} = \sqrt{\frac{(m_{\text{tot}} g)^3}{2 \pi r_p^2 n_p \rho}} \quad \mathrm{[W]},
\end{equation}
where $m_{\text{tot}}$ is the UAV mass, $g$ is the earth gravity, $r_p$ is the propeller radius, $n_p$ is the number of propellers, and $\rho$ is the air density.

The movement-related energy consumption is modeled as:
\begin{equation}
    \label{Energy_move}
    E_t^{\text{move}} = T_{\text{move}} (P_{\text{hover}} + P_{\text{move}}) \quad \mathrm{[J]},
\end{equation}
where $T_{\text{move}} = {\sqrt{(x^u_t-x^u_{t-1})^2 + (y^u_t-y^u_{t-1})^2}}/{v^u}$ is the time needed for the UAV to move from one position to another (at a fixed height), and $P_{\text{move}}$ is the power consumed due to movement calculated as~\cite{8613833}:
\begin{equation}
    \label{move_power}
    P_{\text{move}} = \frac{P_{\text{max}}-P_{\text{idle}}}{v_{\text{max}}} \: v^u + P_{\text{idle}} \quad \mathrm{[W]},
\end{equation}
where $P_{\text{max}}$ and $P_{\text{idle}}$ are hardware-related power values and $v_{\text{max}}$ is the UAV maximum speed. Note that $P_{\text{move}}$ denotes the additional propulsion power above hovering, so the flight power during motion is $P_{\text{hover}} + P_{\text{move}}$.

The total energy consumption at time is simply the sum of both components:
\begin{equation}
    \label{energy_total}
    E_t^{\text{total}} = E_t^{\text{com}} + E_t^{\text{move}} \quad \mathrm{[J]}.
\end{equation}
Similarly, the total time consumed by the UAV to move and communicate with a device can be expressed as:
\begin{equation}
    \label{time_total}
    T_{\text{total}} = T_{\text{com}}^k + T_{\text{move}} + T_{\text{exec}} \quad \mathrm{[s]},
\end{equation}
where $T_{\text{exec}}$ is the inference time of a decision-making algorithm. It is worth noting that the UAV has the option to remain stationary (\textit{i.e.,} hover) and to serve no devices. In that case, we assume that $T_{\text{total}} = T_{\text{exec}}$.

\subsection{Information Freshness}
Age of information (AoI) is a critical metric that quantifies the freshness of information in transmitted packets \cite{Kosta-AoIbook-2017}. For each device $k$, its AoI is estimated as the difference between the current system time and the timestamp of the most recently received packet, effectively reflecting the timeliness of signal updates and state estimation. The individual AoI is calculated as:
\begin{equation}
    \label{AoI}
    A^k_{t+1} =
	\begin{cases}
		0, & \: \text{if device} \ k \ \text{is served}, \\
		\text{min}\{A_{max},A^{k}_t + T_{\text{total}}(t)\}, & \: \text{otherwise}; 
	\end{cases}
\end{equation}
where $A_{max}$ is a predefined upper limit on AoI to ensure bounded modeling. Incorporating AoI alongside energy in the optimization objective is crucial for promoting fairness in user scheduling and maintaining up-to-date signal-state awareness across networked devices~\cite{Kosta-AoIbook-2017,eldeeb2022multi}.

\subsection{Problem Definition}
The objective is to jointly optimize the UAV trajectory and scheduling policy by determining the UAV movement directions $w_t^x,w_t^y$ and the served device $s_t$, so as to minimize both the UAV energy consumption and the average AoI of the devices. The problem is formulated as:
\begin{subequations}\label{P1}
	\begin{alignat}{2}
	\mathbf{P1:}\qquad &\underset{\boldsymbol{w_t^x,w_t^y,s_t}}{\min}       &\ \ \ & \sum_{t=1}^H \Bigg[ \lambda \sum_{k = 1}^{K}\delta^k A^k_t + (1-\lambda) \: E_t^{\text{total}}\Bigg],\label{P1:a}
	\ \\
	&\text{s.t.}   &      & \sum_{t=1}^{H} T_{\text{total}}(t) \leq T_{\text{th}}, \label{P1:b}
	\end{alignat}
\end{subequations}
where $\lambda$ is a weighting factor that balances information freshness and energy efficiency, and $\delta^k \in [0,1]$ is a priority weight for device $k$. The variable $H$ denotes the number of decision steps in an episode. Each step $t$ has a duration $T_{\text{total}}(t)$ in seconds, and an episode terminates once the accumulated physical time exceeds the time budget $T_{\text{th}}$.

Problem \textbf{P1} in \eqref{P1} is NP-hard due to its combinatorial nature and the complex interaction between UAV movement, scheduling, wireless channel dynamics, and coupled signal-processing constraints. Obtaining exact solutions is computationally intractable. Therefore, to address this challenge, we propose a Diffusion-SAC approach that leverages offline RL and diffusion models to learn data-driven, near-optimal signal-control policies from offline data.

\begin{figure*}[t!]
    \centering    \includegraphics[width=1.4\columnwidth,trim={0 5cm 0 4cm},clip]{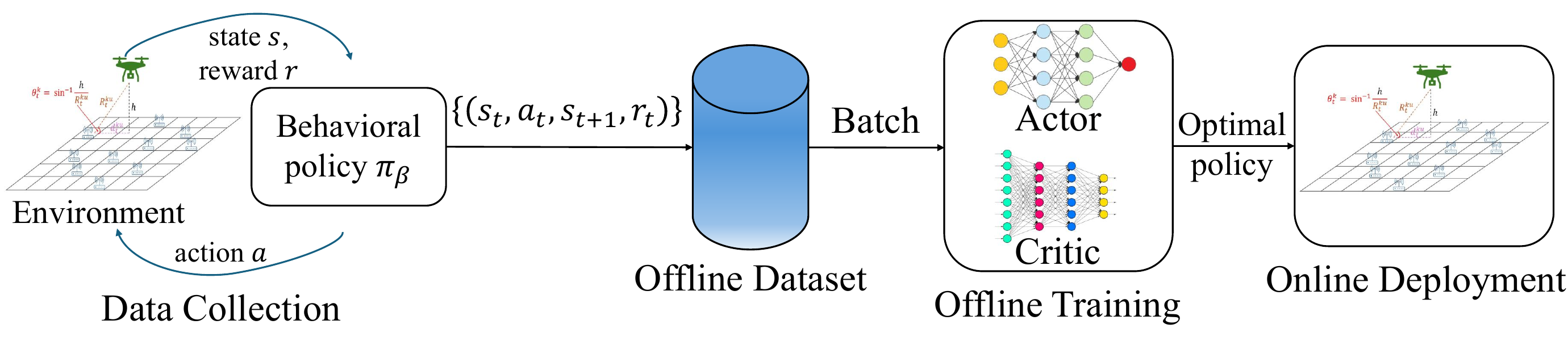} 
    \caption{Offline RL pipeline. An offline dataset $\mathcal{D}$ of state-action-reward tuples is collected from prior interactions and used to train: (i) a state-conditioned diffusion actor $\pi_\phi$ that generates discrete actions via one-hot sampling and categorical projection, and (ii) a critic $Q_\theta$. At deployment, only the learned diffusion actor is used. Given state $s_t$, it runs $N$ denoising steps to propose $x_0$, which is projected to a valid action $a_t$ in the action space.}
    \vspace{0mm}
    \label{Offline_RL}
\end{figure*}

\section{Background}\label{sec:backg} 
In this section, we review the essential concepts of offline RL and diffusion models to introduce the proposed diffusion-based RL algorithm for adaptive signal processing and control in the next section.

\subsection{Online RL}
\label{ssc:online-rl}
Consider a discrete-time environment modeled as a Markov decision process (MDP) defined by the tuple $\langle \mathcal{S}, \mathcal{A}, \mathcal{R}, \mathcal{O}\rangle$, where $\mathcal{S}, \mathcal{A}, \mathcal{R}, \mathcal{O}$ represent, respectively, the sets of state, actions, rewards, and observations. 

At each time step $t \in \{1,2,\cdots\}$, the agent observes a state $s_t \in \mathcal{S}$, takes an action $a_t \in  \mathcal{A}$, transits to a new state $s_{t+1} \in \mathcal{S}$ with transition probability $\Pr(s_{t+1} | s_t,a_t)$ and receives a reward $r_t \in \mathcal{R}$. The objective is to learn a policy $\pi(a|s)$ that maximizes the expected discounted cumulative rewards (or return) 
$J(\pi) = \mathbb{E} \left[\sum_{t=0}^\infty \gamma^t r_t \right]$, 
where $\gamma \in (0,1)$ is the discount factor that controls the weight of future rewards in the objective. Several functions are used to estimate the expected return. 

The state-value function $V(s)$ estimates the expected cumulative reward starting from a state $s$ onward following a specific policy, whereas the state-action value function (Q-function) $Q(s, a)$ estimates the expected cumulative reward starting from state $s$ taking action $a$, and thereafter following policy $\pi$.

Q-learning methods are off-policy, value-based RL algorithms that estimate the optimal Q-function by minimizing the temporal difference (TD) error. In large dimension problems, function approximators are used to model the Q-function or the policy. For example, in deep RL, such as DQNs~\cite{DQNs}, the Q-function is modeled as a neural network. Similarly, actor-critic methods use neural networks to model both the Q-function and the policy. To improve estimation stability in noisy or stochastic environments, experience replay and target Q-networks are proposed. The former saves the agent's experiences during training to enhance the Q-function estimation. The latter decouples the dependencies between the Q-function estimator and the target values, thereby stabilizing training~\cite{DQNs}. The parameters are trained using the following loss:
\begin{align}
\label{bellman_error}
    \mathcal{L}_{\text{q}} = \:& \hat{\mathbb{E}} \left[ \left(  
    \underbrace{r +\gamma \max_{a^{\prime}\in \mathcal{A}} Q_{\phi^{\prime}}(s^{\prime},a^{\prime})}_{\Psi} - Q_{\phi}(s,a) \right)^2 \right],
\end{align}
where $Q_{\phi^{\prime}}$ is the target Q-network parameterized with $\phi^{\prime}$, $Q_{\phi}$ is the Q-network parameterized with $\phi$, and $\Psi$ is the TD error.

\subsection{Offline RL via CQL}
Consider an offline dataset $\mathcal{D}$ composed of the experience tuple $\langle s_t,a_t,r_t,s_{t+1} \rangle$ collected via a behavior policy $\pi_{\beta}$ as in Fig.~\ref{Offline_RL}. In that case, the offline dataset will be used by the agent to optimize the optimal control policy without allowing any online environmental interaction. Training deep RL algorithms directly in an offline manner using a static dataset generally fails due to biased Q-value estimation and overestimation problems~\cite{opt_off_RL}. Such problems arise from a distributional shift between the learned policy and the policies sampled from the dataset. A natural question arises: Why does this issue not occur in off-policy online RL algorithms (\textit{e.g.}, DQNs), even though there is a distribution shift between the learned policy and replay buffer contents? The answer lies in online data collection, which gradually corrects the distribution mismatch. Such an advantage is absent in offline RL~\cite{levine2020offline}.

Conservative Q-learning (CQL) is an offline RL framework that mitigates distributional shift by regularizing the learned Q-function to remain conservative on out-of-distribution (OOD) actions, thereby improving estimation stability. It is easy to build on top of existing deep RL frameworks, such as SAC, which is an off-policy maximum entropy algorithm~\cite{SAC_paper}. In general, actor-critic algorithms combine value-based with policy-based methods. This family of algorithms models the policy using a neural network (actor) that selects the actions, besides modeling the Q-function (critic) that evaluates the quality of the chosen actions~\cite{NIPS1999_6449f44a}. They are particularly suitable for signal processing and control problems, as they enable joint estimation and optimization of decision variables under uncertainty. The critic is updated as in~\eqref{bellman_error} (policy evaluation), while the actor is updated towards an action that maximizes the Q-values (policy improvement).

In CQL, the critic is trained by adding a conservative term to the loss in~\eqref{bellman_error} as follows:
\begin{equation}
 \label{crtici_CQL}
\mathcal{L}_{\text{critic}}\!=\! \frac{1}{2}\mathcal{L}_{\text{q}}\!+\!\alpha \hat{\mathbb{E}} \!\bigg[ \log \sum_{\tilde{a}\in \mathcal{A}}
\exp \bigl( Q_{\phi}(s,\tilde{a}) \bigr) 
    \!-\! \ Q_{\phi}(s,a)  \!\bigg],
\end{equation}
where $\alpha>0$ is a hyperparameter~\cite{kumar2020conservative}. 

{\LinesNumberedHidden
\begin{algorithm}[!t]
\SetAlgoLined

\textbf{Input:} Discount factor $\gamma$, conservative penalty constant $\alpha$, Q-function learning rate $\eta_Q$, policy learning rate $\eta_{\pi}$, number of training iterations $J$, and offline dataset $\mathcal{D}$

\textbf{Initialize:} Actor parameters $\theta$ and critic parameters $\phi$

\For{\text{iteration} $j$ in $\{1$,...,$J$\}}{

\For{\text{batch} $\mathcal{B}$ in $\mathcal{D}$}{

Update the critic parameters using the loss in~\eqref{crtici_CQL}:
$\phi\leftarrow \phi - \eta_Q \: \nabla_{\phi} \: \mathcal{L}_{\text{critic}}$

Update the actor parameters using the loss in~\eqref{actor_CQL}:
$\theta \leftarrow \theta - \eta_{\pi} \: \nabla_{\theta} \: \mathcal{L}_{\text{actor}}$

}
}

\textbf{Return} Optimized Q-function $Q_{\phi}(s,a)$ and policy $\pi_{\theta}(a|s)$ 
\caption{Conservative Q-learning for Offline RL}
\label{CQL_Alg}  \vspace{0mm}
\end{algorithm}}
\vspace{-1mm}

The actor is trained via entropy regularization loss:
\begin{equation}
    \label{actor_CQL}
    \mathcal{L}_{\text{actor}}=\: - \mathbb{E} \bigg[Q_{\phi}(s,a) - \log \: \pi_{\theta}(a|s)\bigg],
\end{equation}
where $\pi_{\theta}$ is the policy modeled as a neural network with parameters $\theta$.  
\algo{CQL_Alg} summarizes the CQL algorithm on top of the SAC architecture.

\subsection{Denoising Diffusion Probabilistic Model} \label{ssc:ddpm}

Consider $x^0$ be the target probability distribution of the input. We define the sequence $x^{M:0}$ as a Markov chain with learned Gaussian transitions with $M$ steps. During the forward process, Gaussian noise is added at each time step to obtain $x^1, x^2, \cdots, x^M$. The relation between two consecutive samples $x^{i-1}$ and $x^i$ is described using the transition probability:
\begin{equation}
    \label{forward}
    q(x^i|x^{i-1}) 
    \sim
    \mathcal{N}(
    \sqrt{1-\beta^i}\, x^{i-1},\beta^i \mathbf{I}),
\end{equation}
where $\mathcal{N}(\mathbf{\mu},\mathbf{\Sigma})$ denotes the normal distribution with mean $\mathbf{\mu} = \sqrt{1-\beta^i} x^{i-1}$, and covariance $\mathbf{\Sigma} = \beta^i \mathbf{I}$, where $\mathbf{I}$ is the identity matrix. Following~\cite{NEURIPS2020_4c5bcfec}, a simple linear schedule is often used for $\beta^i$, though improved schedules such as the cosine schedule have also been proposed~\cite{nichol2021improved}.

Given the Markovian property that the current sample only depends on the previous one, the distribution of $x^M$ given $x^0$ is given by:
\begin{equation}
    \label{forw_dist}
    q(x^M|x^0) = \prod_{i=1}^M q(x^i|x^{i-1}).
\end{equation}
A direct mapping from $x^M$ to $x^0$ is:
\begin{equation}
    \label{mapping}
    x^i = \sqrt{\bar{\alpha}^i} x^0 + \sqrt{1-\bar{\alpha}^i}\,\epsilon,
\end{equation}
where $\alpha^i = 1-\beta^i$, $\bar{\alpha}^i = \prod_{k=1}^i \alpha^k$ and $\epsilon \sim \mathcal{N}(0,\mathbf{I})$. As we increase $i$, the sample becomes pure noise following a normal distribution.

\begin{figure}[t!]
    \centering    \includegraphics[width=1
    \columnwidth,trim={0 0 0 0},clip]{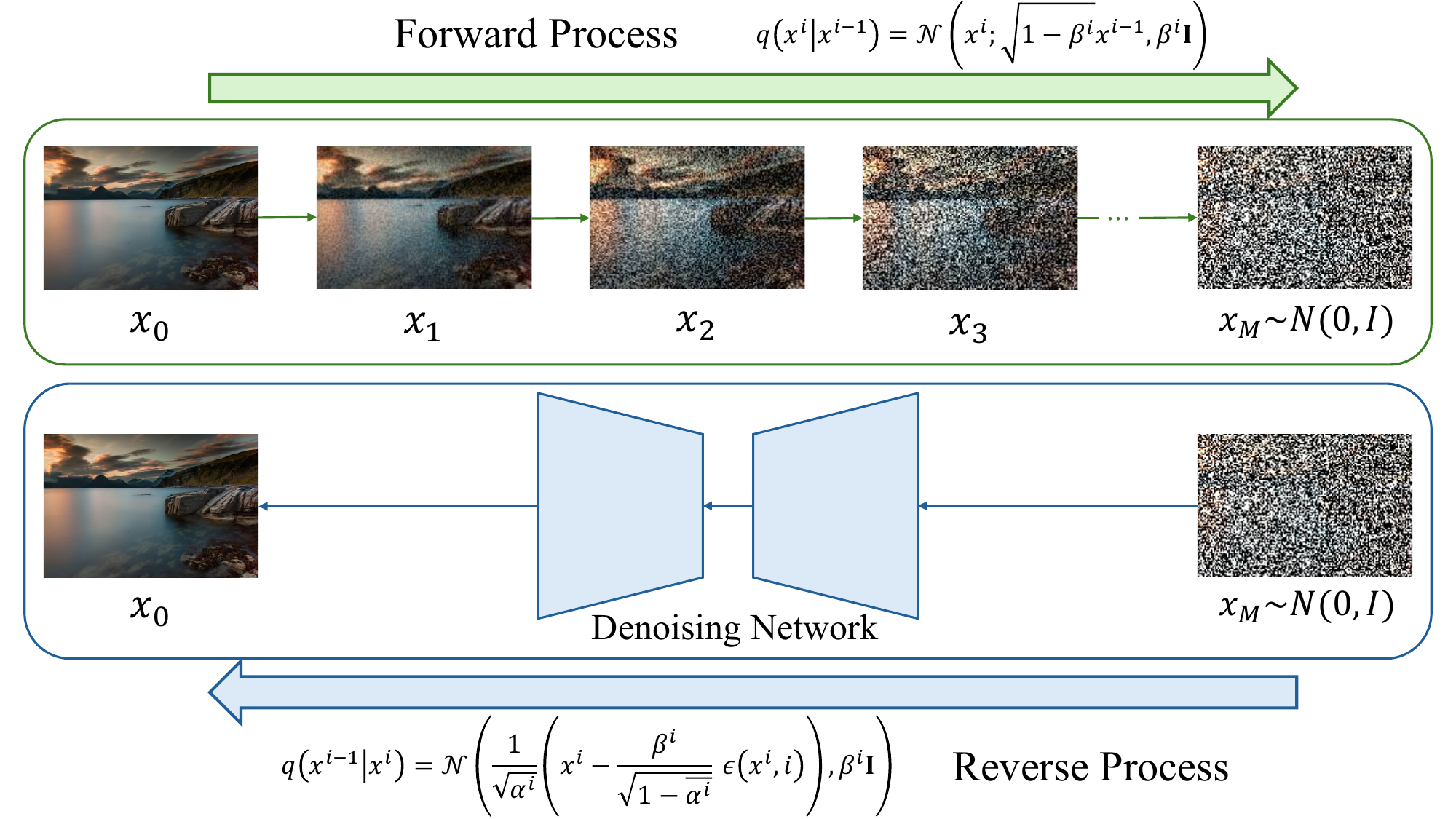} 
    \caption{Illustration of the diffusion policy. \textit{(top)} Forward process: a clean one-hot action vector $x_0$ is gradually perturbed through $q(x_i \!\mid\! x_{i-1})$ up to $x_M$. \textit{(bottom)} Reverse process: conditioned on the current state $s$, the actor parameterization $p_\phi(x_{i-1}\!\mid\! x_i,s)$ iteratively reconstructs a categorical action representation. The final $x_0$ is projected to a discrete action. The state-conditioning implements guidance toward high-value actions while staying close to the offline data.}
    \vspace{0mm}
    \label{diff_illustr}
\end{figure}

In the reverse process (sampling), the goal is to reconstruct $x^0$ from a noisy sample by removing the noise. The transition probability from $x^{i-1}$ given $x^i$ can be constructed using Bayes' theorem, where it follows a Gaussian distribution:
\begin{equation}
    \label{reverse}
    q(x^{i-1}|x^i, x^0) \sim \mathcal{N}\Bigg(\frac{1}{\sqrt{\alpha^i}}\bigg(x^i-\frac{\beta^i}{\sqrt{1-\bar{\alpha}^i}}\epsilon (x^i,i)\bigg),\tilde{\beta}^i \mathbf{I}\Bigg),
\end{equation}
where the variance is
\[
    \tilde{\beta}^i = \frac{1-\bar{\alpha}^{i-1}}{1-\bar{\alpha}^i}\,\beta^i.
\]

This construction enables obtaining $x^0$ from pure noise; however, the noise is now $\epsilon(x^i, i)$, thus dependent on the sample and step. Hence, \cite{NEURIPS2020_4c5bcfec} models the noise as a parameterized neural network, that is,  $\epsilon_{\theta} (x^i, i)$, to predict the noise. In practice, the model is trained using a simplified evidence lower bound (ELBO) objective, which reduces to a mean squared error (MSE) between true and predicted noise:
\begin{equation}
\label{diff_loss}
    \mathcal{L}(\theta) = \mathbb{E}_{x^0,\epsilon,i} \left[\left|\left| \epsilon - \epsilon_{\theta} \left(\sqrt{\bar{\alpha}^i}x^0 + \sqrt{1 - \bar{\alpha}^i} \epsilon, i\right) \right|\right|^2\right],
\end{equation}
where $\epsilon$ is sampled as $\mathcal{N}(0,\mathbf{I})$. Fig.~\ref{diff_illustr} emphasizes the forward and backward processes of the DDPM algorithm.

\section{Diffusion-SAC for Offline UAV Optimization}\label{sec:ODRL}
This section presents the proposed diffusion soft actor-critic (Diffusion-SAC) algorithm for offline RL in UAV-enabled systems. We begin by formulating the problem as an MDP, followed by a detailed explanation of the Diffusion-SAC framework.

\subsection{MDP Formulation}
We model the UAV trajectory and scheduling problem as an MDP as described in Section~\ref{ssc:online-rl}. We assume an episodic environment that terminates after a predefined number of time steps. Bearing this in mind, the MDP of the optimization problem is formulated as follows:
\subsubsection{State space}
The state space consists of three elements $s_t = [x_t^u, y_t^u, A_t^1, \dots, A_t^K, E_t^{\text{total}}]$, where the first two elements correspond to the projected coordinate of the UAV, $A_t^1, \dots, A_t^K$ are the individual AoI of the $K$ devices, and $E_t^{\text{total}}$ is the total consumed energy at time $t$. This composite state captures both spatial and signal-level system information and serves as the sufficient statistic for decision-making.

\subsubsection{Action space}
The action space consists of three elements $a_t = [w^x_t,w^y_t,s_t]$, where $[w^x_t,w^y_t]$ are the movement distances of the UAV and $s_t \in \{0,1,\dots,K\}$ is the served device. The actions $(w^x_t,w^y_t)$ are continuous, while the action $s_t$ is discrete. Therefore, we formulate a hybrid discrete-continuous RL problem, typical in joint communication-control optimization. Note that $(w^x_t,w^y_t) = (0,0)$ means that the UAV hovers without moving, while $s_t = 0$ means that the UAV serves no devices.

\subsubsection{Reward}
The instantaneous reward is a weighted combination of the average AoI and the UAV energy consumption:
\begin{equation}
    \label{reward}
    r_t = -\frac{\lambda}{T} \sum_{k=1}^K \delta^k A_t^k - \frac{1 - \lambda}{T} E_t^{\text{total}},
\end{equation}
where the negative sign arises from the target of minimizing these terms, thereby maximizing the reward. This design naturally balances temporal information freshness with physical-layer energy efficiency. 

\subsubsection{State transition probability}
The state transition probability affects the three elements of the state space. The UAV position is updated as $(x^u_{t},y^u_{t}) + (w^x_t,w^y_t)$. To enforce boundary constraints, a mirroring mechanism is applied: if the next position exceeds the grid limits, the UAV remains at its current location. The AoI values and energy consumption are updated using \eqref{AoI} and~\eqref{energy_total}, respectively.

\subsubsection{Normalization}
To enhance training stability and generalization, and to avoid large deviations between the values of the state and reward elements, we perform state and reward normalization. This procedure stabilizes training and generalizes learning across different data points, especially in offline training~\cite{reward_norm}. Thus, the state and reward are normalized as follows:
\begin{align}
    \bar{s} &= \frac{s-\mu_s}{\sigma_s + \varepsilon}, \label{state_norm} \\
    \bar{r} &= \frac{r-\mu_r}{\sigma_r + \varepsilon}, \label{rewards_norm}
\end{align}
where $\mu_s$ and $\mu_r$ denote the means, $\sigma_s$ and $\sigma_r$ the standard deviations, and $\varepsilon$ a small constant to avoid division by zero. This normalization ensures stable signal statistics for the learning process.

\subsection{Diffusion RL}
Diffusion models are used to model the policy (i.e., the actor) due to their flexible generative parameterization. This is particularly advantageous in offline RL. 

Following Section~\ref{ssc:ddpm}, the actor is defined as a conditional denoising diffusion model. Starting with noise $a^M \sim \mathcal{N}(0,1)$, the reverse process is applied as follows:
\begin{align}
    \label{reverse_QL}
    a^{i-1}|a^i \!&=\! \frac{a^i}{\sqrt{\alpha^i}} \!-\! \frac{\beta^i}{\sqrt{\alpha^i(1\!-\!\bar{\alpha_i})}} \epsilon_{\theta}(a^i,i,s) \!+\! \sqrt{\beta_i} \epsilon,\\
    & \text{for~}  i \!=\! M, \dots, 1, \nonumber
\end{align}
where $\epsilon_{\theta}(a^i,i,s)$ is the denoising network, which generates denoising samples conditioned on the state $s$. We apply the sample rule with $\epsilon = 0$ when $i = 1$. This generative process serves as a stochastic signal reconstruction mechanism, translating noisy policy samples into feasible UAV control actions.

The actor loss in~\eqref{actor_CQL} can be replaced by a diffusion loss in~\eqref{diff_loss} whose target is to reconstruct the action given the state. The actor loss can be re-formulated as follows:
\begin{align}
    \label{diffRL_loss}
    \mathcal{L}_{\text{bc}}\!(\!\theta) \!=\! \mathbb{E}_{i,\epsilon,(s,a)\sim \mathcal{D}}\! 
    \left[\!\left|\!\left| \epsilon \!-\! \epsilon_{\theta} 
    \left(\sqrt{\bar{\alpha}^i}a \!+\! \sqrt{1 \!-\! \bar{\alpha}^i} \epsilon,i,s\right) \right|\!\right|^2\!\right]\!\!,
\end{align}
where $\mathcal{L}_{\text{bc}}$ is behavior cloning (BC) loss, which imitates the learning in the dataset. 

\begin{figure}[t!]
    \centering    \includegraphics[width=1\columnwidth,trim={0 0 0 0},clip]{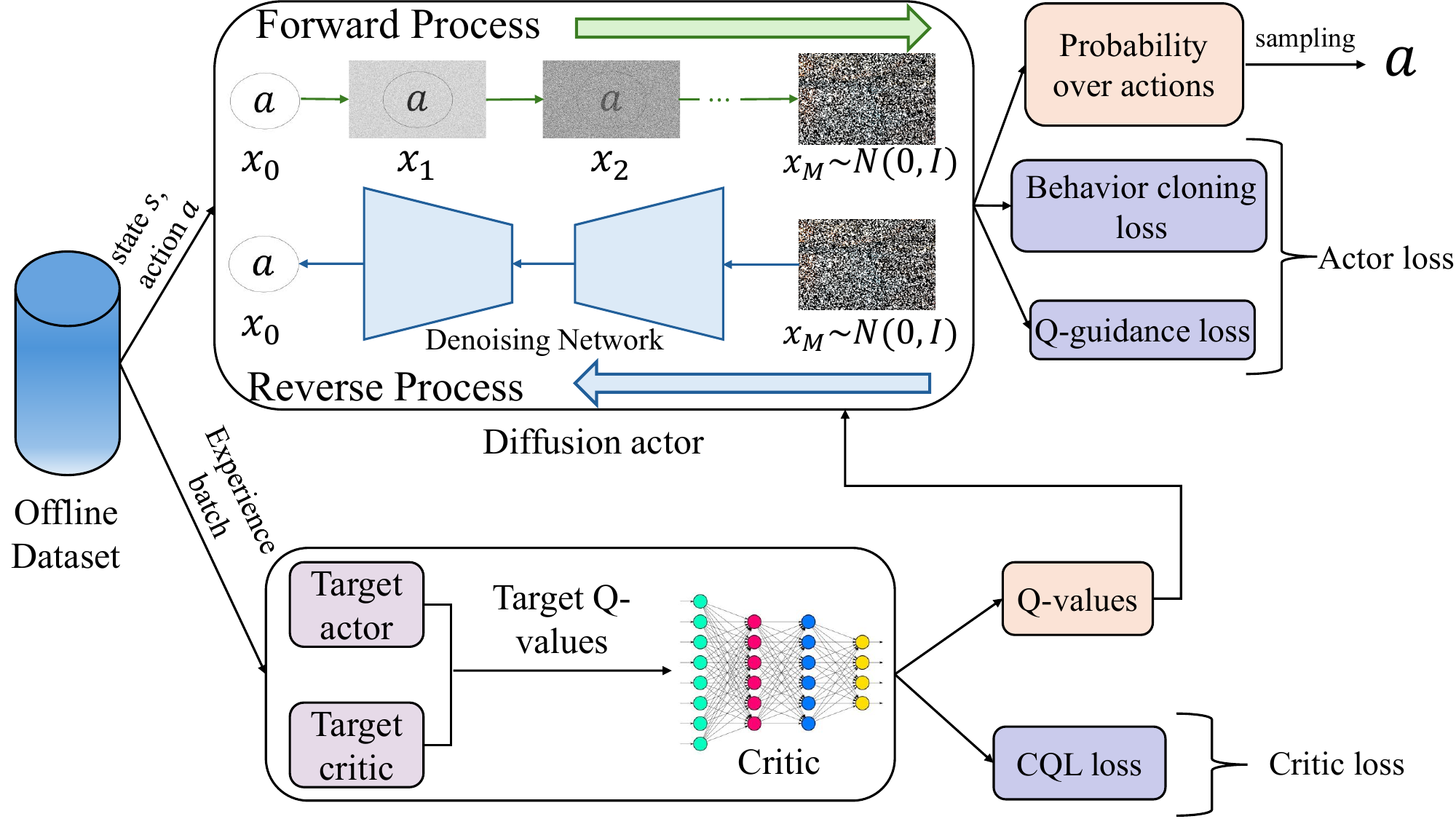} 
    \caption{An illustration of the proposed Diffusion-SAC for offline UAV policy planning. A batch is sampled to compute the CQL loss, which is used to update the critic using the target actor and target critic networks. The critic produces Q-values that the actor uses to compute the Q-guidance loss. The policy (actor) is modeled as a diffusion model conditioned by the state to generate the action. The actor is updated using a weighted combination of the behavior cloning loss and the Q-guidance loss.}  
    \label{diff_QL_illustr}
\end{figure}

Training the actor with the BC loss alone merely copies the behavior policies in the dataset, without exploring new policies that could outperform the known ones. To go beyond imitation, we introduce a Q-guidance loss~\cite{wang2023diffusionpoliciesexpressivepolicy}:
\begin{equation}
    \label{Q_guidance_Loss}
    \mathcal{L}_{\text{g}} = - \mathbb{E}_{s \sim \mathcal{D},a \sim \pi_{\theta}}[Q_{\phi}(s,a)], 
\end{equation}
where $Q_{\phi}$ is the parameterized Q-function (critic) following \eqref{bellman_error} and \eqref{crtici_CQL} and $a$ denotes the hybrid action sampled from the diffusion actor conditioned on $s$.

Finally, the actor loss is computed as a weighted combination of both BC loss and Q-function guidance loss:
\begin{equation}
    \label{actor_loss_all}
    \mathcal{L}_{\text{actor}} = \eta \: \mathcal{L}_{\text{bc}} + (1-\eta) \: \mathcal{L}_{\text{g}},
\end{equation}
where $\eta \in [0,1]$ is a hyperparameter that balances imitation and exploration. High $\eta$ values approaching $1$ would reduce RL to imitation learning, in which the actor imitates the samples seen in the dataset. In contrast, lower $\eta$ values, approaching $0$, eliminate the BC, which acts as a regularizer for the actor, and encourage exploration at the cost of stability. This balance parallels the bias-variance trade-off in adaptive signal estimation, thereby controlling generalization to unseen states.

The critic is trained using the conventional CQL loss, similar to~\eqref{crtici_CQL}, while sampling the next action from the diffusion policy network, $a^{\prime 0} \sim \pi_{\theta^{\prime}}$. Setting the conservative parameter $\alpha = 0$ yields the traditional Bellman update rule as in~\eqref{bellman_error}. This is beneficial when the weight of the BC loss is significant, mitigating the need for regularization. In contrast, large $\alpha$ values are needed to constrain the learned policies to be as close as possible to the behavioral policies when the BC loss weight is relatively small. We recall that deep RL requires the implementation of separate target networks for stability. Hence, we build a target actor and a target critic, which are updated separately and softly. Fig~\ref{diff_QL_illustr} summarizes the proposed training procedure of the proposed Diffusion-SAC.

{\LinesNumberedHidden
\begin{algorithm}[!t]
\SetAlgoLined
\textbf{Input:} Discount factor $\gamma$, conservative penalty constant $\alpha$, Q-function learning rate $\eta_Q$, policy learning rate $\eta_{\pi}$, number of training iterations $J$, denoising steps $I$ and offline dataset $\mathcal{D}$

\textbf{Initialize:} Actor parameters $\theta$ and critic parameters $\phi$

\For{\text{iteration} $j$ in $\{1$,...,$J$\}}{

\For{\text{batch} $\mathcal{B}$ in $\mathcal{D}$}{

Sample $a^{\prime 0}$ using~\eqref{reverse_QL}

Update the critic parameters using the loss in~\eqref{crtici_CQL}:
$\phi\leftarrow \phi - \eta_Q \: \nabla_{\phi} \: \mathcal{L}_{\text{critic}}$

Sample $a^{0}$ using~\eqref{reverse_QL}

Update the actor parameters using the loss in~\eqref{actor_loss_all}:
$\theta \leftarrow \theta - \eta_{\pi} \: \nabla_{\theta} \: \mathcal{L}_{\text{actor}}$
}
}

\textbf{Return} Optimized Q-function $Q_{\phi}(s,a)$ and policy $\pi_{\theta}(a|s)$ 
\caption{Diffusion-SAC for offline RL}
\label{DSAC_alg}  \vspace{0mm}
\end{algorithm}}

The original diffusion Q-learning work in~\cite{wang2023diffusionpoliciesexpressivepolicy} was developed exclusively for continuous action spaces. In this work, we extend the framework to hybrid discrete-continuous control. Specifically, the action is represented as a single continuous vector composed of two normalized displacement components and a one-hot encoding of the discrete scheduling decision. During diffusion training, the full hybrid vector is treated as a continuous signal and optimized using the standard DDPM noise-prediction objective. At inference time, the continuous components are rescaled to physical units, while the discrete segment is interpreted as logits and converted to a categorical action via softmax followed by argmax. The critic receives normalized continuous actions and one-hot discrete encodings, enabling consistent evaluation across both modalities. Since the diffusion actor does not provide an explicit likelihood, no entropy term is included; instead, stochasticity arises from the denoising sampling process. Algorithm~\ref{DSAC_alg} summarizes the proposed Diffusion-SAC for offline RL.

\section{Experimental Analysis}\label{sec:results}
In this section, we present the experimental analysis of the proposed approach. First, we describe the simulation setup, data collection, and pre-processing procedures. We then present extensive and diverse numerical results to validate the effectiveness of our method from both control and signal-processing perspectives.

\subsection{Simulation Setup}
We consider a UAV operating in a $1000 \times 1000$ square meters area, serving $10$ randomly deployed sensors. The UAV flies at a fixed altitude of $100$ meters with a constant velocity of $25$ m/s. At the beginning of each episode, the UAV's initial position is randomly initialized to ensure diverse spatial conditions and channel realizations. The state space comprises $13$ elements: $2$ elements representing the UAV's 2D position, $10$ elements representing the individual AoI (Age of Information) of each device, and $1$ element representing the total energy consumption. The action space consists of $11$ possible discrete actions derived from device selection choices (including the idle/no-service case) and $2$ continuous actions derived from the Cartesian movement directions.

If a device is scheduled for service, it transmits a data packet of size $2$ Mb to the UAV. The transmission time, AoI evolution, and energy consumption are functions of the achievable data rate, which is determined by the channel condition. Both the actor and critic networks consist of three hidden layers, each with $256$ neurons. The Adam optimizer is employed to update network parameters, and training is performed using mini-batches of size $64$. All experiments are tested over $100$ independent test environments with randomized device layouts and propagation conditions. All reported performance curves correspond to the mean over these 100 environments. Simulations are implemented using PyTorch and executed on a single NVIDIA Tesla V100 GPU. The complete list of simulation parameters used in the experiments is summarized in Table~\ref{UAV_Parameters}. Our open-source code is available at \url{https://github.com/Eslam211/Diffusion_UAV}.

\subsection{Baseline Methods and QoS Metrics}
First, we compare the proposed algorithm to conventional offline RL algorithms, such as conservative Q-learning (CQL)~\cite{kumar2020conservative}, Batch-Constrained Deep Q-Learning (BCQ)~\cite{fujimoto2019offpolicydeepreinforcementlearning}, and implicit Q-learning (IQL)~\cite{kostrikov2021offlinereinforcementlearningimplicit}. In addition, we perform an ablation study by adjusting the values of $\alpha$ and $\eta$ in the proposed methods, \textit{i.e.,} changing the weights of CQL loss, behavior cloning loss, and Q-guidance loss in~\eqref{crtici_CQL} and~\eqref{actor_loss_all}. We also include comparisons with heuristic-based trajectory and scheduling policies widely adopted in the literature~\cite{eldeeb2022multi}:
\begin{itemize}
    \item Time-division multiplexing (TDM): Devices are served in a round-robin manner at equal time intervals.

    \item Saver: The UAV prioritizes serving the closest device to minimize energy usage.

    \item Random-walk (RW): The UAV selects actions randomly.
\end{itemize}

\begin{table}[t!]
\centering
\caption{Simulation parameters and hyperparameters~\cite{ghorbel2019joint,ghazzai2017energy,9507262}} 
\label{UAV_Parameters}
\begin{tabular}{cc|cc}
\toprule
\textbf{Parameter}                                    & \textbf{Value} & \textbf{Parameter}                                    & \textbf{Value} \\ \midrule
\midrule

$R_c$ & $100$ m & $v^u = v_{\text{max}}$ & $25$ m/s\\
$h$ & $100$ m & $f_c$ & $2$ GHz\\
$C$ & $10$ & $D$ & $0.6$\\
$\xi^{\text{LoS}}$ & $1$ dB & $\xi^{\text{NLoS}}$ & $20$ dB\\
$v^l$ & $3 \times 10^8$ m/s & $B$ & $1$ MHz \\
$N_0$ & $-174$ dBm / Hz & $D_k$ & $2$ Mb \\
$m_{\text{tot}}$ & $0.5$ Kg & $g$ & $9.807$ m/$\text{s}^2$ \\
$r_p$ & $0.2$ m & $n_p$ & $4$ \\
$\rho$ & $1.225$ Kg / $\text{m}^3$ & $T_{\text{exec}}$ & $0.02$ s \\
$P_{\text{max}}$ & $5$ W & $P_{\text{idle}}$ & $0$ W \\
$P_{\text{com}}$ & $0.0126$ W & $T = A_{\max}$ & $400$ \\
$\lambda$ & $0.5$ & $\delta^k$ & $\frac{1}{K}$ \\
$E_{\text{threshold}}$ & $3500$ J & $\gamma$ & $0.99$ \\
Batch size & $64$ & Optimizer & Adam \\
$M$ & $50$ & Iterations $J$ & $150$ \\

\bottomrule
\end{tabular} 
\end{table}

\begin{figure*}[t!]
    \centering
    \subfloat[Return\label{online_rew}]{\includegraphics[width=0.6875\columnwidth]{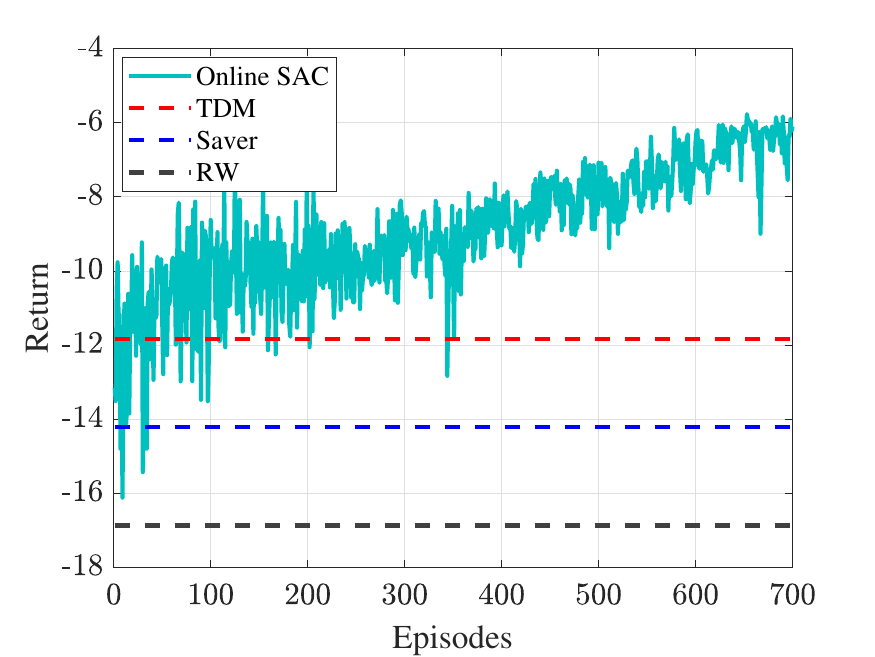}}
    \subfloat[AoI\label{online_AoI}]{\includegraphics[width=0.6875\columnwidth]{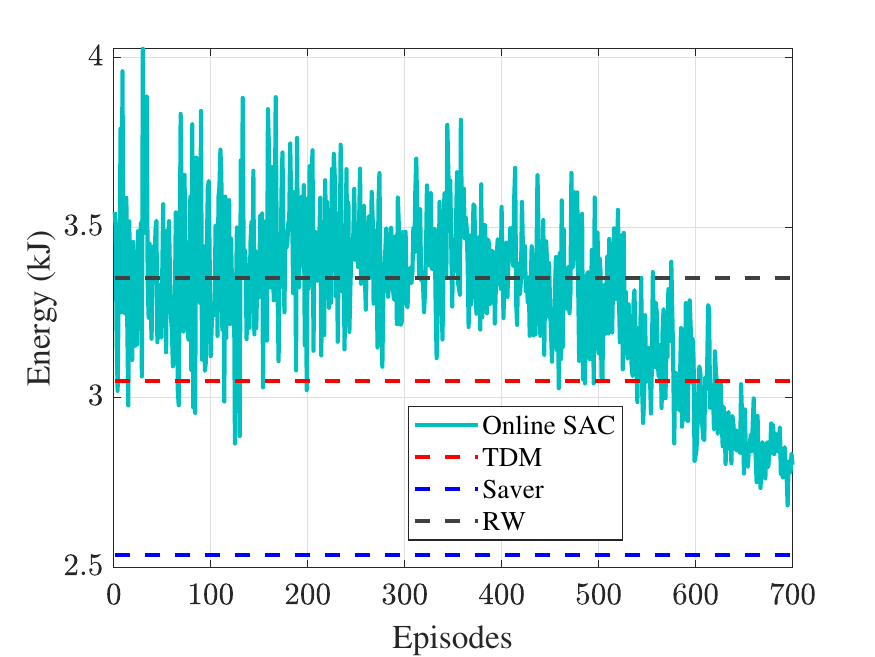}}
    \subfloat[Energy\label{online_energy}]{\includegraphics[width=0.6875\columnwidth]{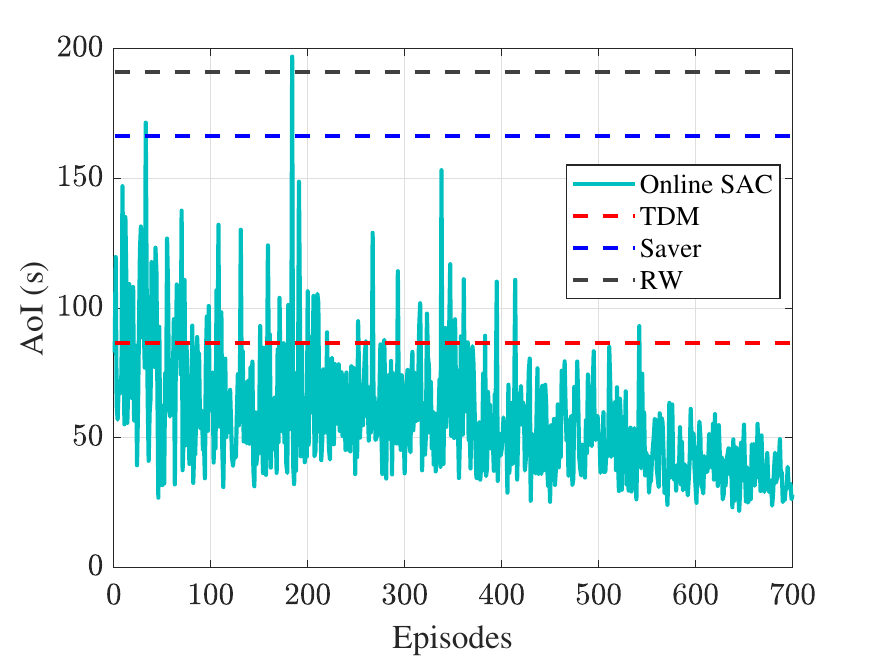}}
    \vspace{-1mm}
    \caption{Performance of online SAC compared to the heuristic baselines in terms of: (a) normalised return, (b) average AoI, and (c) energy consumption.}
    \vspace{-2mm}
    \label{Online_perf}
\end{figure*}

\begin{figure*}[t!]
    \centering
    \subfloat[Return\label{diff_rew}]{\includegraphics[width=0.6875\columnwidth]{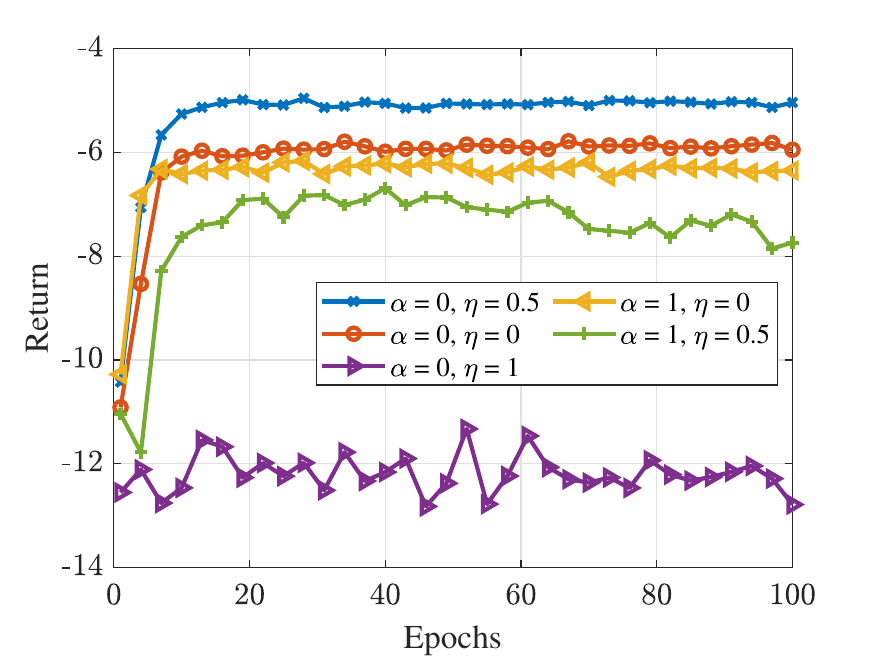}}
    \subfloat[AoI\label{diff_AoI}]{\includegraphics[width=0.6875\columnwidth]{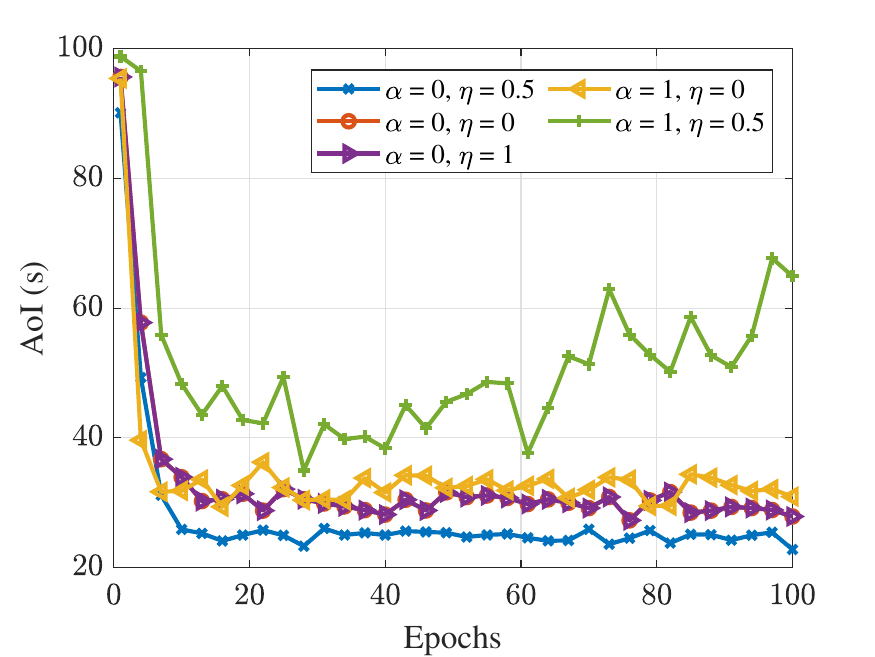}}
    \subfloat[Energy\label{diff_energy}]{\includegraphics[width=0.6875\columnwidth]{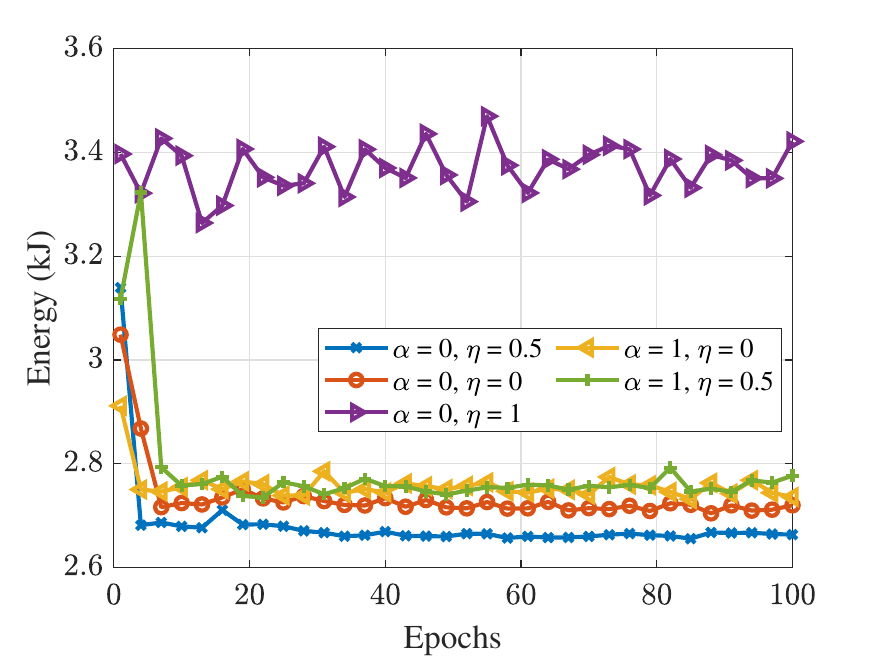}}
    \vspace{-1mm}
    \caption{Performance of different variants of the proposed algorithm in terms of: (a) normalised return, (b) average AoI, and (c) energy consumption.}
    \vspace{-2mm}
    \label{diff_only}
\end{figure*}

The quality-of-service (QoS) metrics include the set of achieved rewards, the average AoI, and the total energy consumption. In addition to these metrics, which are implicitly captured by the reward function, we also report the total throughput and the average packet transmission time, reflecting physical-layer and temporal efficiency. Together, these metrics provide a comprehensive view of system performance from both communication and control standpoints. We further demonstrate the impact of the number of denoising steps, dataset size, and dataset quality on the convergence of the proposed algorithm. Additionally, we analyze the computational complexity of the proposed model as we adjust the number of denoising steps, and compare these results with conventional offline RL algorithms to highlight the inherent trade-offs between efficiency and complexity in diffusion-based signal modeling.

\subsection{Data Collection and Online RL}
In the first experiment, illustrated in Fig.~\ref{Online_perf}, we present the online training performance of an online SAC agent. The agent is trained over $700$ episodes using both real-time interactions with the environment and sampled experiences from a replay buffer. As shown in the figure, the SAC agent exhibits steady convergence and significantly outperforms traditional baselines, including TDM, Saver, and RW. As the cumulative reward increases, both the average AoI and the total energy consumption decrease, indicating improved communication efficiency and signal freshness through adaptive trajectory and scheduling control. This result demonstrates the potential of learning-based methods to optimize UAV-assisted data-collection systems by reducing communication latency and energy expenditure.

Despite its strong performance, online RL methods face critical limitations, particularly in safety-critical and resource-constrained environments. The need for extensive online exploration can be costly, risky, and impractical in real-world UAV deployments. Such exploration can also distort the signal environment or violate QoS constraints, limiting real-time applicability. To address this, we turn to offline RL techniques, which eliminate the need for live interactions during training by leveraging pre-collected datasets.

To construct the offline dataset, we use the replay buffer collected during online SAC training. Data gathered in the early stages of training are generally of lower quality, reflecting suboptimal decisions and exploratory behavior. In contrast, data collected in later stages correspond to more refined, near-optimal policies. This variation in data quality plays an important role in shaping the learning dynamics and final performance of the offline RL algorithms evaluated in subsequent experiments. The final dataset contains approximately $30$k transition tuples sampled uniformly across the entire training horizon, thereby incorporating both early exploratory trajectories and later near-optimal behaviors. This mixture promotes diverse state-action coverage rather than restricting the dataset to a single deterministic behavior policy.

\begin{figure*}[t!]
    \centering
    \subfloat[Return\label{baselines_rew}]{\includegraphics[width=0.6875\columnwidth]{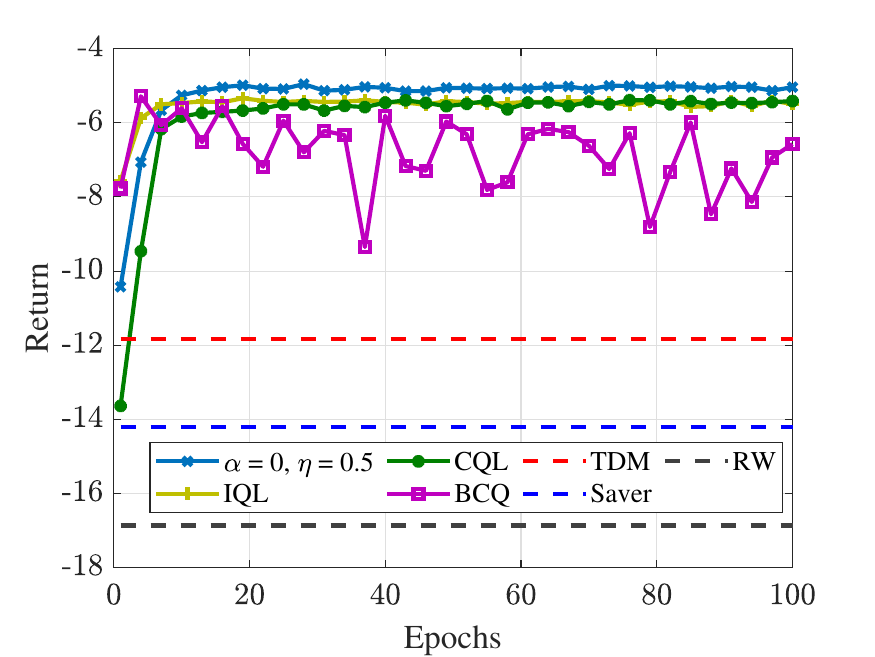}}
    \subfloat[AoI\label{baselines_AoI}]{\includegraphics[width=0.6875\columnwidth]{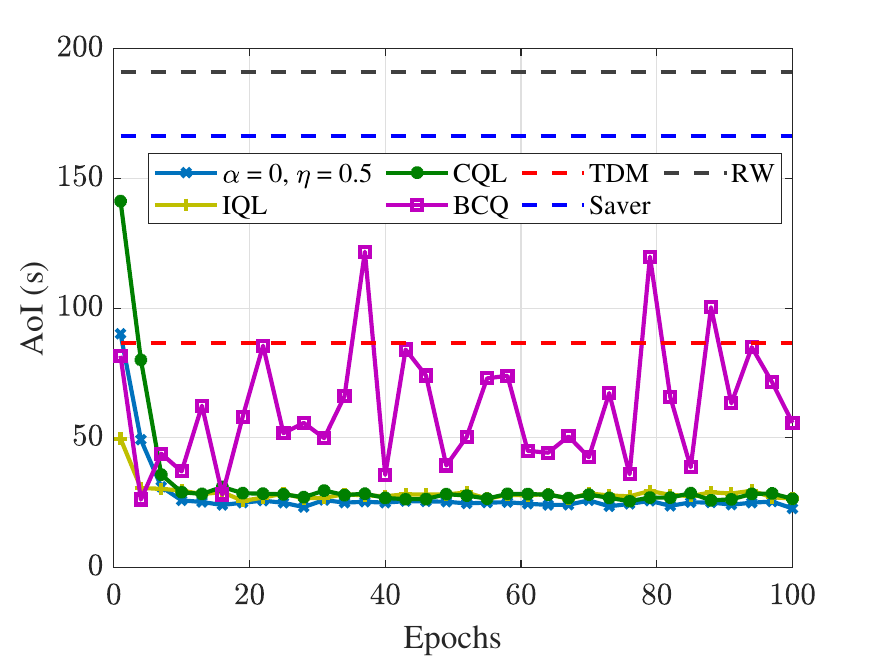}}
    \subfloat[Energy\label{baselines_energy}]{\includegraphics[width=0.6875\columnwidth]{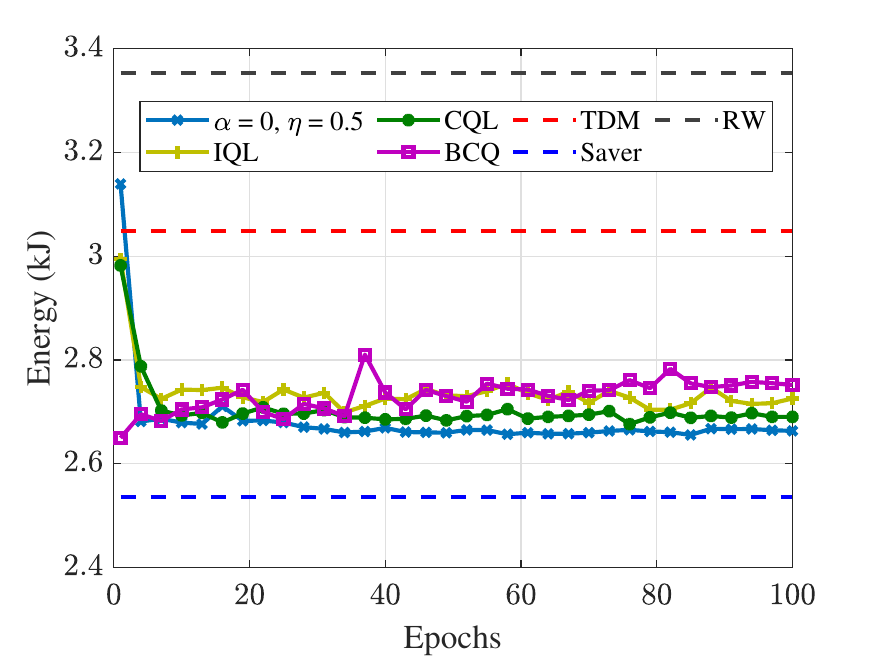}}
    \vspace{-1mm}
    \caption{Performance of the proposed algorithm compared to other offline baselines in terms of: (a) normalised return, (b) average AoI, and (c) energy consumption.}
    \vspace{-2mm}
    \label{Baselines}
\end{figure*}

\begin{figure*}[t!]
    \centering
    \subfloat[Dataset size\label{Rewards_dataset_size}]{\includegraphics[width=0.6875\columnwidth]{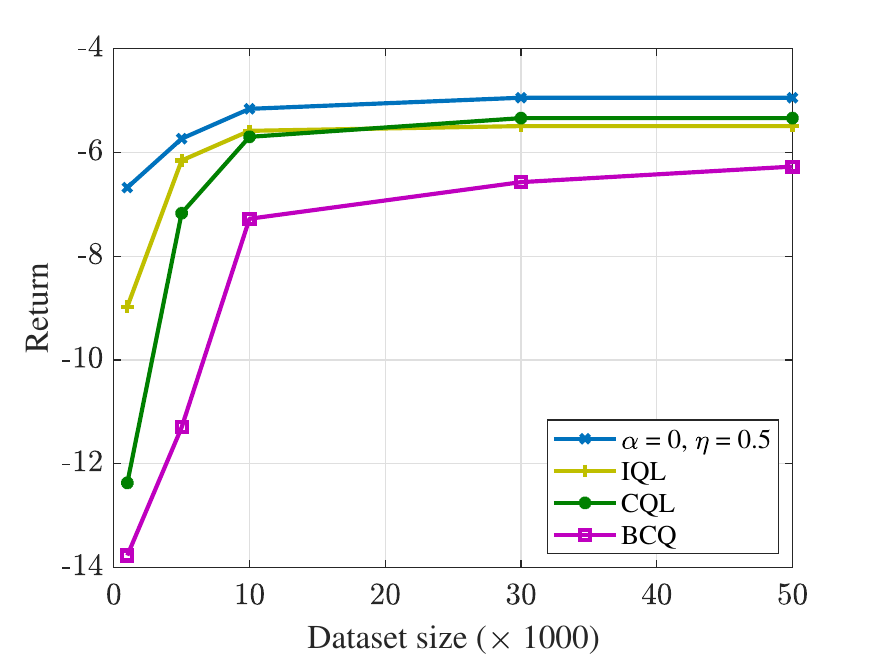}}
    \subfloat[Good dataset\label{Good_dataset}]{\includegraphics[width=0.6875\columnwidth]{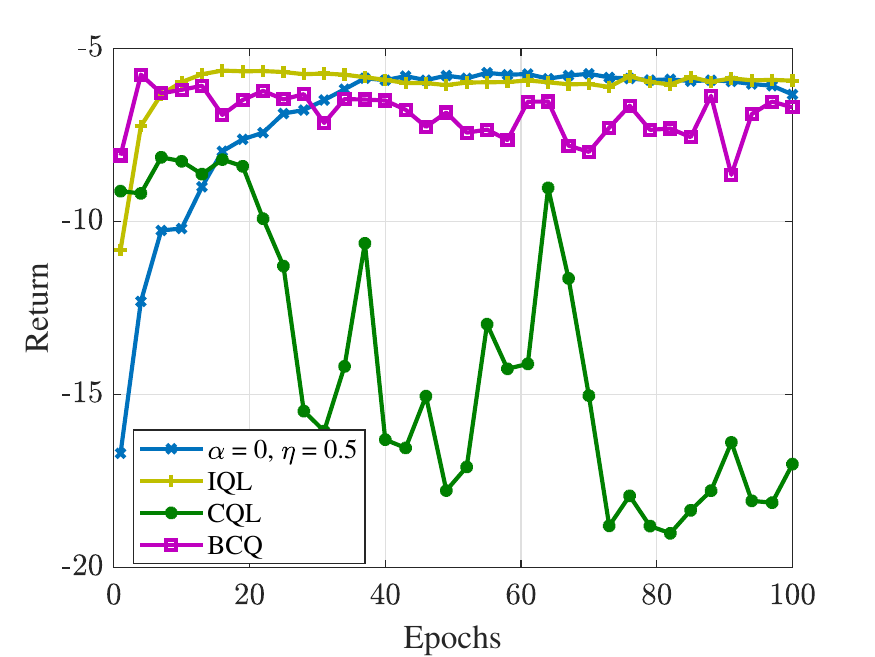}}
    \subfloat[Bad dataset\label{Bad_dataset}]{\includegraphics[width=0.6875\columnwidth]{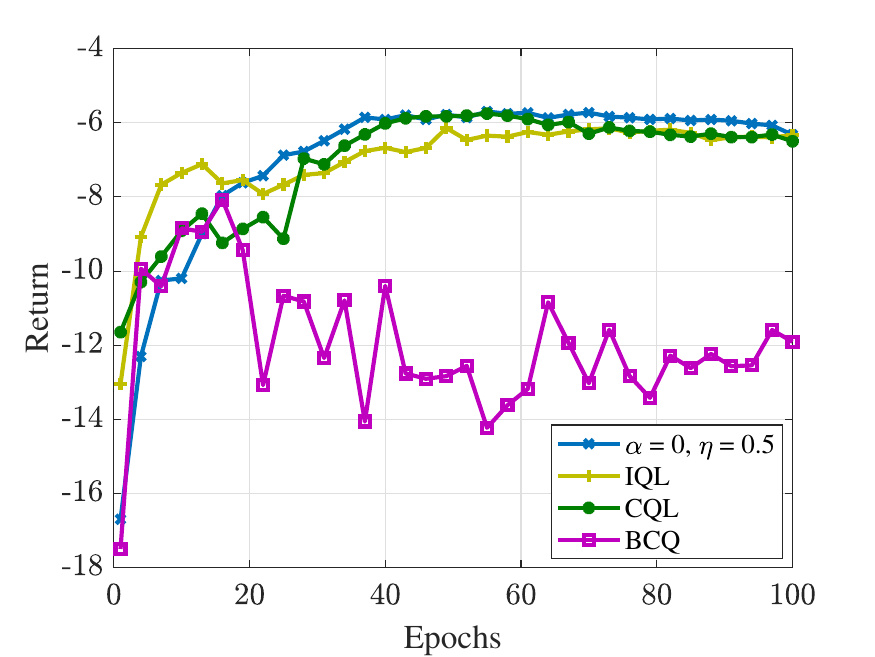}}
    \vspace{-1mm}
    \caption{An illustration of the effect of the dataset size and dataset quality on the proposed algorithm compared to the baselines.}
    \vspace{-2mm}
    \label{Datasets}
\end{figure*}

\subsection{Numerical Results}
In the next experiment presented in Fig.~\ref{diff_only}, we evaluate the performance of the proposed Diffusion-SAC algorithm under varying configurations of the loss function parameters introduced in~\eqref{crtici_CQL} and~\eqref{actor_loss_all}. Specifically, we analyze the impact of different combinations of the CQL loss weight ($\alpha$) and the actor loss weighting parameter ($\eta$), which balances the BC and Q-guidance terms. Fig.~\ref{diff_rew} depicts the average normalized return for each configuration. Setting $\alpha = 0$ effectively disables the CQL loss, while adjusting $\eta$ modifies the influence of BC and Q-guidance losses. For instance, $\eta = 0$ corresponds to a pure Q-guidance objective (no BC), whereas $\eta = 1$ removes the Q-guidance term entirely, reducing the actor loss to standard behavior cloning.

Among the tested variants, the configuration with $\alpha = 0$ and $\eta = 0.5$, \textit{i.e.}, removing the CQL loss while maintaining a balance between BC and Q-guidance, yields the highest return. This suggests that behavior cloning serves as a sufficient constraint on OOD actions, effectively mitigating distributional shift without the conservatism introduced by CQL. Interestingly, even when both CQL and BC losses are omitted (\textit{i.e.}, $\alpha = 0$, $\eta = 0$), the model still achieves competitive performance, demonstrating the ability of diffusion models to implicitly learn structured signal representations and policy dynamics through Q-guidance alone.

In contrast, including the CQL loss ($\alpha = 1$) tends to over-constrain the policy, regardless of whether BC is present, leading to convergence to sub-optimal solutions. This overconservatism limits the model's capacity to generalize or improve upon the dataset's behavior. The case where $\alpha = 0$ and $\eta = 1$, which removes both CQL and Q-guidance losses, effectively reduces the learning to pure imitation. While this setup may perform adequately on high-quality datasets, it is less robust when the dataset includes suboptimal or noisy trajectories. Finally, Figs.\ref{diff_AoI} and\ref{diff_energy} validate these findings by showing that the Diffusion-SAC configuration with $\alpha = 0$ and $\eta = 0.5$ also achieves superior performance in minimizing both AoI and energy consumption, confirming its effectiveness in optimizing multi-objective UAV communication tasks.

In the experiments presented in Fig.~\ref{Baselines}, we compare the performance of the proposed Diffusion-SAC algorithm against several state-of-the-art offline RL baselines. For clarity and conciseness, we include only the best-performing configuration of Diffusion-SAC (\textit{i.e.}, $\alpha = 0$ and $\eta = 0.5$) as established in previous experiments. A high-quality offline dataset of $30$k samples, collected to ensure broad coverage of the state-action space, is used for training all algorithms.

As illustrated in Fig.~\ref{baselines_rew}, the BCQ algorithm yields the lowest average return, indicating its limited ability to generalize from the dataset in this environment. Both IQL and CQL perform comparably, achieving relatively high and stable returns, yet falling short of the optimal policy. In contrast, the proposed Diffusion-SAC consistently achieves the highest return, indicating its superior capacity to leverage the dataset while avoiding over-conservatism or poor generalization.

This trend is further corroborated in Fig.\ref{baselines_AoI}, where Diffusion-SAC achieves the lowest average AoI, slightly outperforming both CQL and IQL. Moreover, as shown in Fig.\ref{baselines_energy}, it also results in the lowest energy consumption among all methods. These results collectively highlight the effectiveness of Diffusion-SAC in optimizing both communication freshness (AoI) and energy efficiency, which are two critical metrics in UAV-assisted data collection, while also surpassing existing offline RL methods in overall performance. The weighting parameter $\lambda$ controls the trade-off between AoI and energy; varying $\lambda$ shifts this balance but does not alter the relative superiority of Diffusion-SAC over baseline methods.

In this set of experiments, we investigate the impact of dataset size and quality on the performance of the proposed algorithm relative to standard offline RL baselines. Fig.~\ref{Rewards_dataset_size} illustrates the effect of varying dataset sizes on policy performance. The proposed method, specifically Diffusion-SAC with $\alpha = 0$ and $\eta = 0.5$, demonstrates remarkable robustness to dataset size. Notably, its performance with just $10$k data samples remains nearly identical to that with $50$k samples, indicating efficient policy learning even from relatively small datasets. In contrast, the baseline algorithms, including IQL, CQL, and BCQ, exhibit significant performance degradation when trained on smaller datasets, highlighting their sensitivity to data volume.

\begin{figure}[!t]
    \centering
    \subfloat[Transmission time \label{Transmission_time}]{\includegraphics[width=0.6875\columnwidth]{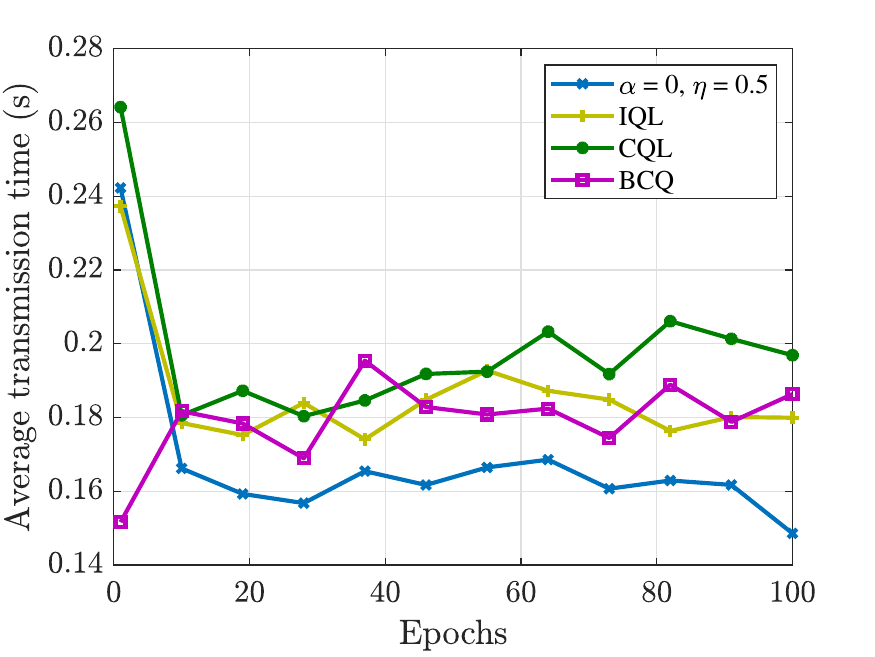}}\\
    \hskip -2.8ex
    \subfloat[Total data \label{Total_data}]{\includegraphics[width=0.6875\columnwidth]{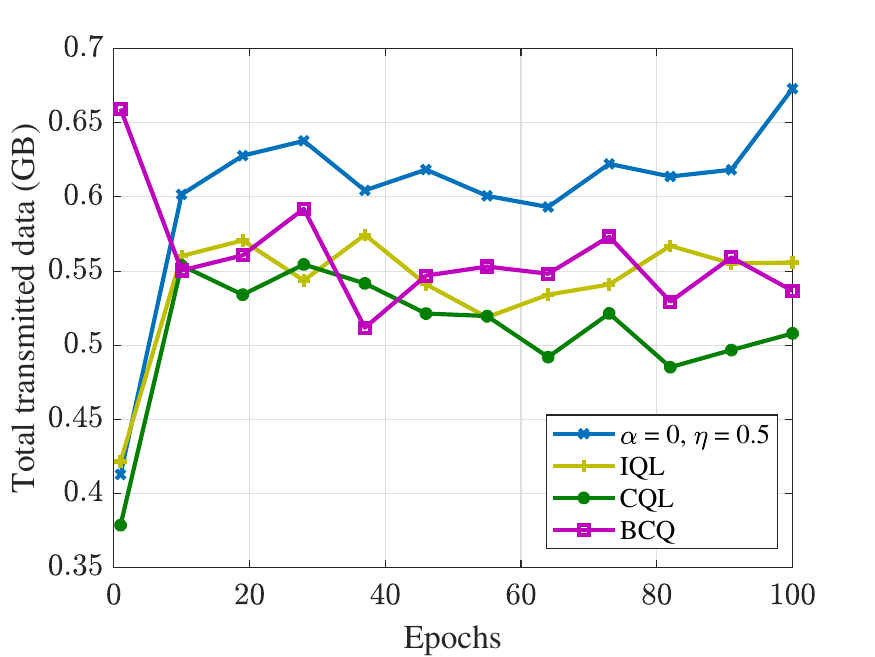}}
    \caption{The performance of the proposed algorithm compared to the baselines in terms of the transmission rate: (a) Average transmission time of one packet and (b) Total transmitted data in one episode.}
    \label{Rates} 
\end{figure}

A similar trend is observed when evaluating the effect of dataset quality, as depicted in Fig.\ref{Good_dataset} and Fig.\ref{Bad_dataset}. When trained on a poor-quality dataset (containing mostly sub-optimal or random behaviors), CQL demonstrates relatively strong performance due to its conservative nature. However, it fails to effectively exploit high-quality datasets, often converging to overly cautious policies. On the other hand, BCQ performs better when the dataset consists predominantly of high-quality trajectories but struggles significantly with noisy or sub-optimal data. IQL shows more balanced behavior across varying dataset qualities, but still underperforms the optimal policies.

The robustness of Diffusion-SAC under reduced dataset size and varying data quality indicates that the learned policy does not merely memorize trajectories but generalizes beyond the dataset. Its stable performance under small or mixed-quality datasets suggests improved resilience to distributional shift compared to conventional offline RL baselines.


In Fig.\ref{Transmission_time}, we present the average transmission time required for a device to send a packet to the UAV successfully. This transmission time is primarily determined by the achievable transmission rate, which in turn depends on the SNR and the UAV's relative position to the transmitter. The proposed Diffusion-SAC algorithm achieves an average transmission time of approximately $0.15$ seconds, outperforming all baseline methods.

This improvement in transmission efficiency is further reflected in Fig.\ref{Total_data}, which shows the total data collected by the UAV over a single episode. The proposed algorithm enables the UAV to collect over $0.65$ GB of data, representing a $35\%$ increase in throughput compared to baseline algorithms. These results underscore the multifaceted benefits of the proposed approach. In addition to effectively minimizing AoI and energy consumption, the Diffusion-SAC also enhances communication efficiency by implicitly maximizing data throughput, thereby demonstrating its overall superiority in optimizing UAV-assisted data collection tasks.

\begin{figure}[!t]
    \centering
    \subfloat[Return \label{Rewards_denoising_steps}]{\includegraphics[width=0.6875\columnwidth]{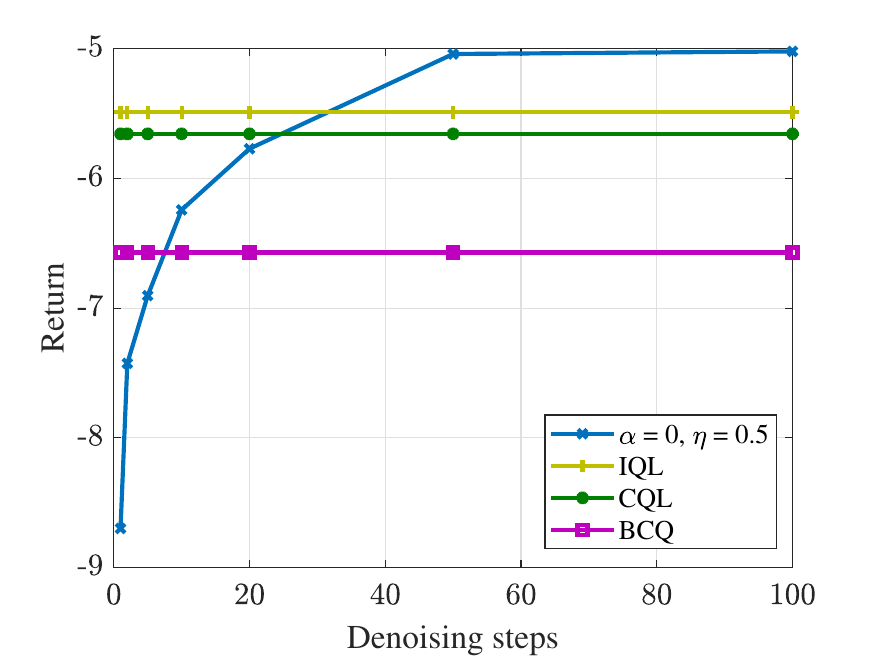}}\\
    \hskip -2.8ex
    \subfloat[Inference Time\label{Inference_Time}]{\includegraphics[width=0.6875\columnwidth]{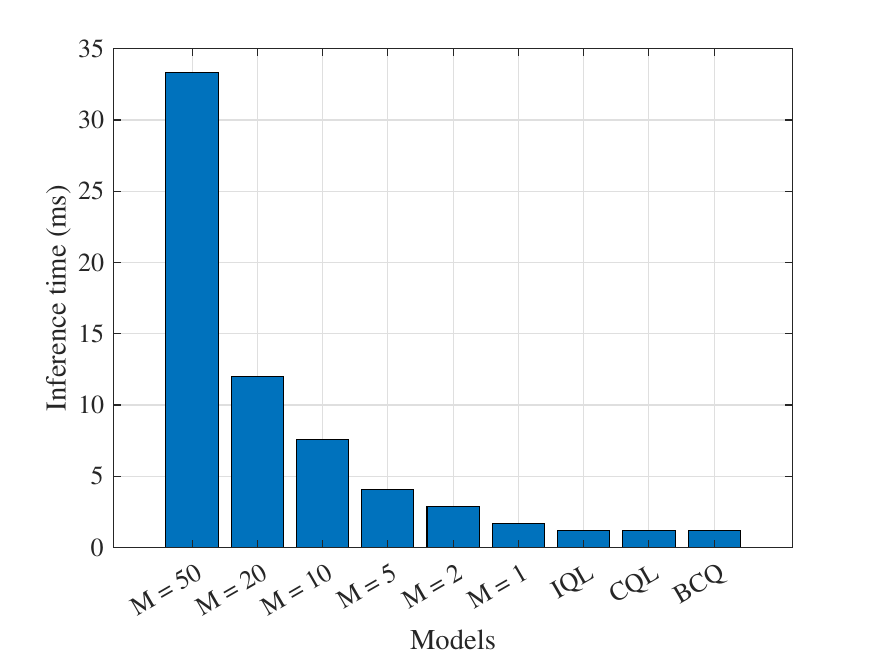}}
    \caption{The return and inference time of the proposed algorithm while changing the number of denoising steps compared to the baselines.}
    \label{Denoising_steps} 
\end{figure}

Finally, we analyze the impact of the number of denoising steps on the performance of the proposed Diffusion-SAC algorithm, as shown in Fig.~\ref{Rewards_denoising_steps}. The results indicate that increasing the number of denoising steps improves convergence, as measured by accumulated rewards. Specifically, a small number of denoising steps (fewer than $10$) yields suboptimal performance, failing to surpass the returns achieved by the baseline algorithms. In contrast, employing more than $20$ denoising steps consistently enables the model to reach near-optimal policies, thereby outperforming all baselines.

To further understand the trade-off, we evaluate the inference time of the proposed method as a function of the number of denoising steps and compare it with the baselines, as shown in Fig.~\ref{Inference_Time}. Since all algorithms utilize the same underlying neural network architecture, the baselines, namely IQL, CQL, and BCQ, achieve minimal inference times of approximately $2$ milliseconds. However, in Diffusion-SAC, inference time scales linearly with the number of denoising steps. For instance, achieving optimal performance with more than $20$ denoising steps incurs an inference delay of around $12$ milliseconds.

Although Diffusion-SAC introduces a moderate increase in inference time, this additional computational cost is justified by its significant gains in core signal-processing performance metrics, including reduced transmission delay, improved energy efficiency, and overall system throughput. From a real-world system perspective, this represents a favorable trade-off between complexity and performance, where over prolonged operation, the benefits in latency reduction and extended device lifetime outweigh the overhead, yielding more reliable, energy-aware signal control in UAV networks.

hile Diffusion-SAC demonstrates consistent performance across randomized environments, performance may degrade when the offline dataset is severely biased or when the number of denoising steps is too small, as illustrated in the numerical results. In such cases, the policy may converge toward conservative or imitation-like behavior. This highlights the importance of sufficient dataset diversity and adequate diffusion depth.

\section{Conclusions}\label{sec:conclusions} 

We developed a Diffusion-SAC algorithm for offline RL. By integrating the generative denoising capabilities of diffusion models with the robust estimation of the CQL algorithm, the proposed framework enables efficient and safe offline policy optimization. The diffusion-based actor models express policy distributions that enhance generalization beyond the dataset behavior while maintaining stability through conservative regularization. We compared different variants of the proposed algorithm by varying the weights of the CQL loss, behavior-cloning loss, and Q-guidance loss, which control the trade-off among imitation, exploration, and conservatism. This analysis identified stable signal-processing operating points that balance data fidelity and decision adaptivity. 

The proposed method was evaluated in a UAV-assisted wireless network scenario, aiming to jointly minimize AoI and energy consumption while ensuring fairness among devices via weighted AoI optimization. Compared to state-of-the-art offline RL algorithms, including CQL, IQL, and BCQ, the proposed Diffusion-SAC algorithm consistently outperforms all baselines, achieving over $35 \%$ higher throughput while minimizing both AoI and total energy consumption. Furthermore, it exhibits stronger resilience to variations in dataset size and quality, demonstrating robust signal generalization under data limitations.

Finally, future work will focus on reducing time complexity and inference latency by minimizing the number of denoising steps using advanced diffusion acceleration techniques, such as the denoising diffusion implicit model (DDIM)~\cite{song2022denoisingdiffusionimplicitmodels}. Extending this framework to multi-agent and distributed signal processing scenarios in next-generation wireless networks also represents a promising direction for future research.

\bibliographystyle{IEEEtran}
\bibliography{IEEEabrv,references}
\end{document}